\documentclass[12pt,authoryear]{elsarticle}
\usepackage{hyperref}

\usepackage{amsmath}
\usepackage{indentfirst}
\usepackage{makecell}
\usepackage{amssymb}
\usepackage{geometry}
\usepackage{graphicx}
\usepackage{float}
\usepackage{subfigure}
\usepackage{color}

\usepackage{multirow}
\usepackage{caption}
\usepackage{booktabs}

\usepackage{algorithm}
\usepackage{algorithmic}
\hypersetup{hidelinks}
\begin{document}

\begin{frontmatter}

\title{Domain-Incremental Remote Sensing Change Detection via Difference-Guided Adaptation and Frequency-Decoupled Distillation}

\author{Daifeng Peng\corref{mycorrespondingauthor}}

\cortext[mycorrespondingauthor]{Corresponding author at: School of Remote Sensing and Geomatics Engineering, Nanjing University of Information Science and Technology, Nanjing
210044, China.}

\makeatletter \emailauthor{daifeng@nuist.edu.cn}{D. Peng}

\author{Yaning Li}
\author{Haiyan Guan}

\address{School of Remote Sensing and Geomatics Engineering, Nanjing University of Information Science and Technology, Nanjing
210044, China}

\begin{abstract}

Remote sensing change detection (RSCD) models are prone to catastrophic forgetting when incrementally adapted to new domains. Existing domain-incremental learning (DIL) methods mainly preserve image-level representations but often overlook bitemporal discrepancy cues, which are critical for robust change detection under domain shifts. To address this limitation, we propose DG-FDD, a domain-incremental change detection framework that integrates Difference-Guided Adaptation and Frequency-Decoupled Distillation. Specifically, the Difference-Guided Dynamic Adapter (DGDA) models bitemporal feature discrepancies to promote change-aware feature adaptation and reduce domain-specific interference. Meanwhile, the Frequency-Decoupled Knowledge Distillation strategy with Cross-domain Synthesis (FDKD-CS) separates structural information from domain style in the frequency domain, enabling stable knowledge transfer without historical data. Extensive experiments on three public high-resolution RSCD datasets under two- and three-domain incremental protocols demonstrate that DG-FDD effectively mitigates catastrophic forgetting. Compared with independently trained single-task models, DG-FDD records mean relative changes in F1 and IoU of only $-0.23\%$ and $-0.45\%$, respectively, across six two-domain sequences, and $-0.69\%$ and $-1.31\%$, respectively, across the three evaluated three-domain sequences. These results indicate a favorable stability--plasticity balance between historical knowledge retention and new-domain adaptation in continual cross-domain change detection. Code will be available at \url{https://github.com/pandaielise/DG-FDD}.

\end{abstract}

\begin{keyword}

Remote sensing change detection; Domain incremental learning; Catastrophic forgetting; Bitemporal discrepancy modeling; Frequency-domain distillation; Knowledge distillation

\end{keyword}

\end{frontmatter}

\section{Introduction}
Remote sensing change detection (RSCD) aims to identify dynamic surface changes by analyzing multi-temporal images acquired over the same geographical area and has become a fundamental technique in applications such as urbanization monitoring~\citep{RN1}, land-use analysis~\citep{RN4}, disaster assessment~\citep{RN5}, and environmental monitoring~\citep{RN6}. In recent years, deep learning-based approaches, including Convolutional Neural Networks (CNNs)~\citep{RN74}, Transformers~\citep{RN9}, and Mamba architectures~\citep{RN10}, have substantially advanced CD performance. Benefiting from their powerful nonlinear modeling capability and hierarchical feature representation learning ability, these methods have substantially improved both the accuracy and automation of CD systems~\citep{RN75}. However, most CD models assume that training and test data are independently and identically distributed, which rarely holds in real-world remote sensing applications. Variations in sensor configurations, imaging conditions, and seasonal characteristics often introduce substantial domain shifts across datasets, thereby limiting the generalization ability of CD models in practical cross-domain scenarios.

Directly fine-tuning a model on newly acquired data typically drives its parameters toward the optimum of the new task, which can substantially degrade its ability to recognize change patterns from previously learned scenes. This phenomenon is commonly known as catastrophic forgetting~\citep{RN18}. Although retraining with all historical data can mitigate this issue, it is often impractical in large-scale scenarios due to storage, computational, and privacy constraints. Therefore, balancing new-domain adaptation with prior-knowledge preservation under limited access to historical data remains a key challenge for the continual deployment of CD models~\citep{RN14}. To address catastrophic forgetting, several mainstream incremental learning paradigms have been developed, including replay-based methods~\citep{RN17}, regularization-based methods~\citep{RN20}, and parameter-isolation methods~\citep{RN25,RN24}. These methods balance historical knowledge preservation and new-task adaptation through experience replay, knowledge regularization, or task-specific parameter decoupling, and have shown promising performance in general visual recognition tasks. Recent advances in large-scale pretrained models have further promoted parameter-efficient continual learning through feature disentanglement, feature synthesis~\citep{RN3}, and visual prompt learning~\citep{RN2}, providing new opportunities to address the stability-plasticity dilemma under limited data availability.

\begin{figure*}[!h]
	\centering
	\includegraphics[width=0.9\textwidth]{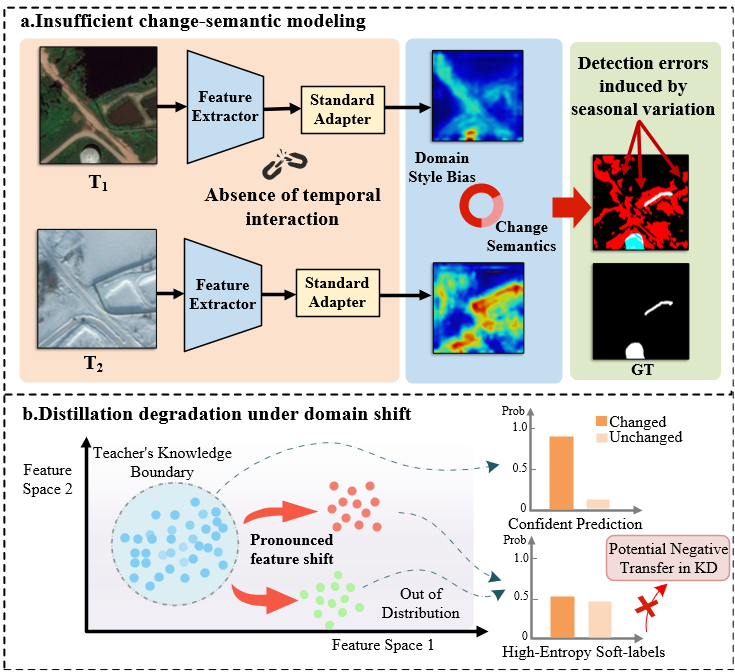}
	\caption{Limitations of existing domain-incremental change detection methods in cross-domain scenarios.}
	\label{fig:fig1}
\end{figure*}
\par
However, directly applying these conventional approaches to domain-incremental change detection (DICD) remains highly challenging. Unlike conventional incremental learning, DICD maintains a fixed label space while facing continuously evolving feature distributions caused by cross-domain variations in remote sensing imaging conditions. This setting gives rise to a central challenge for high-resolution RSCD under continual deployment: preserving previously learned change-structure knowledge with limited access to historical data while adapting to new-domain imaging characteristics induced by geographic, seasonal, and sensor variations. The mismatch between stable semantic supervision and drifting feature distributions can induce gradient conflicts during optimization, resulting in knowledge confusion and severe forgetting. MDINet~\citep{RN27},  a representative DICD method, partially mitigates catastrophic forgetting through domain-specific residual adapters and hierarchical knowledge distillation. Nevertheless, existing DICD methods for high-resolution remote sensing imagery still have two key limitations. First, most parameter-efficient fine-tuning modules are inherited from single-temporal vision tasks and process bitemporal images through separate branches, as shown in Fig. 1(a). This weakens their ability to exploit temporal comparison cues and may cause overfitting to domain-specific image styles rather than robust cross-domain change semantics. Second, existing distillation and feature alignment strategies generally assume consistent teacher--student feature distributions. In cross-domain remote sensing scenarios, however, substantial source--target distribution shifts are common, as illustrated in Fig. 1(b). These shifts can lead to biased teacher responses and semantic inconsistency during direct feature alignment, thereby degrading knowledge transfer in continual cross-domain learning.

To address these challenges, this paper proposes a DICD framework for remote sensing imagery that integrates a Difference-Guided Dynamic Adapter (DGDA) and frequency-domain distillation. Unlike existing methods that independently optimize temporal features or directly distill teacher representations, the proposed framework jointly exploits bitemporal discrepancy cues in the spatial domain and structural knowledge in the frequency domain. Specifically, DGDA incorporates bitemporal feature discrepancies into parameter adaptation, enabling change-relevant semantics to guide feature recalibration and suppress domain-specific appearance interference. To improve cross-domain knowledge transfer, we further develop a Frequency-Decoupled Knowledge Distillation with Cross-Domain Feature Synthesis (FDKD-CS) strategy, which enforces phase-spectrum consistency and introduces cross-domain synthesized features during distillation. This design preserves structural knowledge while reducing the adverse effects of domain discrepancy on teacher supervision, thereby achieving a favorable balance between plasticity and stability without relying on historical data replay.
The main contributions of this paper are summarized as follows.

\par
\begin{itemize}
  \item[$\bullet$]We propose DG-FDD, a DICD framework that jointly integrates difference-guided adaptation and frequency-decoupled knowledge distillation to mitigate catastrophic forgetting under continuous domain shifts.
   \item[$\bullet$]We design a Difference-Guided Dynamic Adapter (DGDA) that exploits bitemporal feature discrepancies to guide dynamic parameter adaptation, thereby enhancing change-relevant semantics while suppressing domain-specific appearance interference.We introduce FDKD-CS, a frequency-decoupled distillation strategy that combines phase-spectrum consistency with amplitude-guided cross-domain feature synthesis to reduce teacher student distribution mismatch during continual adaptation.
  \item[$\bullet$]Extensive experiments are conducted on three public high-resolution RSCD datasets under two-task and three-task domain-incremental protocols, demonstrating that  DG-FDD achieves a favorable balance between stability and plasticity in
  continual cross-domain change detection.

\end{itemize}

\section{Related Work}
\subsection{Deep learning-based change detection}
Deep learning has become the dominant paradigm in RSCD due to its strong capability for hierarchical feature learning and nonlinear semantic modeling. Unlike conventional methods that rely on handcrafted features or shallow change representations~\citep{RN28}, end-to-end deep models learn change-discriminative representations directly from bitemporal images, thereby improving the extraction of complex change patterns~\citep{RN30}. Existing RSCD networks have evolved from CNN-based Siamese and encoder-decoder architectures to attention-enhanced models~\citep{RN37,RN38}, Transformer-based models~\citep{RN40,RN39}, CNN-Transformer hybrids~\citep{RN46,RN45}, and recent Mamba-based architectures~\citep{RN48,RN47}. These developments have substantially improved spatial representation, long-range dependency modeling, and computational efficiency. Nevertheless, most existing methods are developed under the assumption of static data distributions. When deployed in open and continuously evolving environments, they remain vulnerable to domain shifts, which can cause severe performance degradation and even catastrophic forgetting.
\subsection{Incremental learning}
Incremental learning (IL) aims to learn new tasks sequentially while retaining previously acquired knowledge. Its central challenge is to balance stability and plasticity. Existing IL methods are commonly grouped into replay-based, regularization-based, and parameter-isolation approaches, with recent work increasingly moving toward hybrid frameworks. Replay-based methods mitigate catastrophic forgetting by rehearsing a small set of historical samples during new-task learning, thereby approximating joint training~\citep{RN49}.To reduce memory costs, recent studies have replaced raw samples with feature representations~\citep{RN50,RN51}, semantic queries~\citep{RN52}, or more effective sample selection strategies~\citep{RN17,RN53}. However, their reliance on stored historical data limits the applicability in memory-constrained or privacy-sensitive scenarios. Regularization-based methods avoid data storage by constraining parameter updates or output responses to preserve prior knowledge~\citep{RN18,RN19}.  Although memory-efficient, they often struggle to maintain an effective stability-plasticity trade-off under distribution shifts. Parameter-isolation methods instead allocate task-specific parameters or lightweight adaptation modules to reduce interference between old and new knowledge~\citep{RN54}. With the rise of pretrained models, parameter-efficient fine-tuning (PEFT) has become increasingly popular in continual learning~\citep{RN57}. By freezing the backbone and introducing lightweight adapters, PEFT enables efficient adaptation with limited parameter overhead~\citep{RN24,RN60}. Recent IL studies have further developed hybrid frameworks that combine experience replay, knowledge distillation, parameter-efficient fine-tuning, and domain adaptation to improve robustness under distribution shifts~\citep{RN49,RN63,RN61}.

In RSCD, IL is further extended to domain-incremental change detection (DICD), where models must continually adapt to new imaging domains caused by sensor, seasonal, and geographic variations. Unlike conventional IL, DICD must preserve historical knowledge while handling continuously drifting feature distributions across domains.. Existing representative methods mainly combine parameter-efficient adaptation with knowledge preservation strategies to alleviate knowledge confusion during continual adaptation~\citep{RN27}. Nevertheless, these methods largely inherit general IL paradigms and insufficiently exploit the unique characteristics of RSCD, particularly bi-temporal discrepancy modeling and robust knowledge transfer under domain shifts.
\subsection{Bi-temporal difference modeling}
CD differs from general remote sensing classification and segmentation because it focuses on identifying semantic changes between two temporally separated observations. Existing CD networks commonly model bitemporal relationships through Siamese feature extraction, feature differencing, temporal attention, and multi-scale fusion~\citep{RN34,RN33}. Recent studies have further improved difference modeling by introducing feature difference maps ~\citep{RN69}, cross-domain feature aggregation strategies~\citep{RN70}, hybrid CNN-Transformer representations~\citep{RN71}, and semantic interaction mechanisms~\citep{RN72}, leading to better discrimination of changed regions and stronger suppression of background interference. However, these methods are mainly developed for one-stage offline training with static data distributions. In DICD, bitemporal features encode not only change semantics but also domain-specific appearance variations caused by sensor, seasonal, and geographic differences. Existing parameter-efficient adaptation methods often treat the two temporal branches independently, leaving bitemporal discrepancies underexploited as guidance for feature adaptation. As a result, domain-specific styles may become entangled with change semantics during continual adaptation, weakening the robustness of cross-domain change representation. Therefore, how to exploit bitemporal discrepancies while disentangling domain-specific styles from change semantics remains a critical yet underexplored challenge in DICD.

\subsection{Knowledge distillation}

Knowledge distillation (KD) is widely used in continual learning to preserve historical knowledge without accessing previous training data. Conventional KD methods mitigate catastrophic forgetting by encouraging the student model to mimic the teacher's output responses~\citep{RN26,RN19}. Recent studies have further extended distillation from prediction logits to intermediate feature representations to retain high-level semantics and fine-grained spatial structures. Typical strategies include multi-level feature alignmen~\citep{RN73,RN25,RN27} and local structural distillation~\citep{RN20}, which have shown effectiveness in dense prediction tasks such as semantic segmentation and object detection.
However, most KD methods assume that the teacher can provide reliable supervision for current-domain inputs. This assumption becomes problematic in DICD, where sensor, seasonal, and geographic variations can cause substantial distribution shifts and biased responses from the frozen teacher. In such cases, directly enforcing feature consistency may introduce semantic mismatch and degrade knowledge transfer, especially for structurally complex change patterns. These limitations motivate distillation strategies that account for teacher reliability under domain shifts.

\section{Methodology}
\subsection{Problem formulation}
\subsubsection{Remote sensing change detection}

Given a pair of strictly registered bitemporal remote sensing images $X = \{X^{t_1}, X^{t_2}\}$, where $X^{t_1}, X^{t_2} \in \mathbb{R}^{H \times W \times C}$ denote remote sensing images acquired at times $t_1$ and $t_2$, respectively. $H$, $W$, and $C$ denote the spatial size and spectral channel number. The goal of RSCD is to learn a parameterized model $\Phi$ to predict the corresponding binary change map $\hat{Y} \in \{0,1\}^{H \times W}$. The value 1 denotes changed regions and 0 denotes unchanged regions.
\par
A CD network $\Phi$ generally consists of a feature extraction encoder $\mathcal{F}_{\theta_{\text{enc}}}$ and a change decoding module $\mathcal{G}_{\theta_{\text{dec}}}$. For the input bitemporal images, the encoder first maps them into a deep feature space and extracts high-level semantic features $F^{t_1} = \mathcal{F}_{\theta_{\text{enc}}}(X^{t_1})$ and $F^{t_2} = \mathcal{F}_{\theta_{\text{enc}}}(X^{t_2})$, where $F \in \mathbb{R}^{H' \times W' \times D}$. The $\mathcal{G}_{\theta_{\text{dec}}}$ then extracts temporal difference relationships between these two feature groups and restores the spatial resolution, finally outputting a change probability map $P \in \mathbb{R}^{H \times W \times 2}$:
\begin{equation}\label{equ-prob_map}
	P = \sigma\left(\mathcal{G}_{\theta_{\text{dec}}}(F^{t_1}, F^{t_2})\right)
\end{equation}
where $\sigma(\cdot)$ denotes the Softmax activation function, $P_c^{(h,w)} \in [0,1]$ denotes the confidence probability predicted by the network that spatial position $(h, w)$ belongs to class $c \in \{0,1\}$.
\par
In conventional static scenarios, the model is optimized end-to-end by minimizing the standard pixel-level Cross-Entropy Loss (CE):
\begin{equation}\label{equ-ce_loss}
	\mathcal{L}_{\text{CE}} = -\frac{1}{H \times W} \sum_{h=1}^{H} \sum_{w=1}^{W} \sum_{c=0}^{1} \mathbb{I}\left(y^{(h,w)} = c\right) \log\left(P_c^{(h,w)}\right)
\end{equation}
where $\mathbb{I}(\cdot)$ denotes the indicator function, which equals 1 when the condition is satisfied and 0 otherwise.
\par
Although this objective effectively learns bitemporal change features, it relies on the assumption of independently and identically distributed training and test data. Under open and dynamic deployment, the mismatch between a fixed label space and drifting feature distributions can undermine static optimization, resulting in knowledge confusion and catastrophic forgetting.
\subsubsection{Domain Incremental Change Detection}
Due to variations in sensor configurations, geographical regions, seasonal conditions, and imaging environments, remote sensing data acquired at different stages often exhibit substantial distribution shifts.
To characterize this dynamic learning process, DICD is formulated as a multi-stage sequential learning task.
Specifically, the model is assumed to experience $T$ incremental stages sequentially, denoted as $t \in \{1, 2, \dots, T\}$.
At the $t$-th stage, only the current-domain dataset is accessible:
\begin{equation}\label{equ-dataset_t}
	\mathcal{D}_t = \left\{(X_{t,i}, Y_{t,i})\right\}_{i=1}^{N_t}
\end{equation}
where $N_t$ denotes the number of samples in the current stage, and $X_{t,i}$ and $Y_{t,i}$ represent the bi-temporal image pair and corresponding binary change label, respectively.
\par
Unlike conventional static learning, DICD involves substantial shifts in the marginal distribution of the input space $\mathcal{X}$ as learning progresses, namely:
\begin{equation}\label{equ-marginal_shift}
	\mathbb{P}(\mathcal{X}_t) \neq \mathbb{P}(\mathcal{X}_{t'}), \quad \forall t \neq t', \quad t, t' \in \{1, 2, \dots, T\}
\end{equation}
where $\mathbb{P}(\mathcal{X}_t)$ denotes the marginal distribution of input images at the $t$-th incremental stage.
\par
Furthermore, different from IL, where the label space continuously expands over time (i.e., $\mathcal{Y}_1 \subset \mathcal{Y}_2 \subset \cdots \subset \mathcal{Y}_T$), the label space in DICD remains unchanged throughout all stages:
\begin{equation}\label{equ-label_space}
	\mathcal{Y}_1 \equiv \mathcal{Y}_2 \equiv \cdots \equiv \mathcal{Y}_T \equiv \{0,1\}
\end{equation}
where $\mathcal{Y}_t$ denotes the label space at the $t$-th incremental stage, $T$ represents the total number of incremental tasks, $\subset$ denotes the proper subset operator, and $\equiv$ indicates strict set equivalence.
\par
Equations~(4) and~(5) highlight the mismatch between a fixed label space and a drifting feature space. Due to domain shifts, identical land-cover categories often exhibit distinct visual characteristics across domains; for instance, a feature pattern indicating real changes in domain $\mathcal{D}_{t-1}$ may merely reflect seasonal variations in domain $\mathcal{D}_t$.

Furthermore, since access to historical data $\mathcal{D}_{1:t-1}$ is restricted at stage $t$, optimizing $\theta_t$ (initialized from $\theta_{t-1}$) solely on $\mathcal{D}_t$ using standard cross-entropy loss biases the model toward the current distribution $\mathbb{P}(\mathcal{X}_t)$. This disrupts the previously established mapping $\mathbb{P}(\mathcal{X}_{1:t-1}) \rightarrow \mathcal{Y}_{1:t-1}$, leading to catastrophic forgetting of historical patterns.

Therefore, the objective of DICD is to learn an optimal parameter set that balances current-domain adaptation and historical knowledge preservation under data-restricted conditions:
\begin{equation}\label{equ-dicd_objective}
	\theta_T^* = \mathop{\arg\min}_{\theta} \sum_{t=1}^{T} \mathbb{E}_{(X_t, Y_t) \sim \mathcal{D}_t} \left[\mathcal{L}\left(\Phi(X_t; \theta), Y_t\right)\right]
\end{equation}
where $\theta_T^*$ denotes the desired global optimal parameter set after completing all $T$ incremental stages, $\mathbb{E}[\cdot]$ represents the expectation operator corresponding to the empirical risk over domain $\mathcal{D}_t$, and $\mathcal{L}(\cdot,\cdot)$ denotes the loss function measuring the discrepancy between predictions and ground-truth labels.
\par
Existing DICD methods typically mitigate catastrophic forgetting through parameter isolation and knowledge distillation. However, parameter isolation often underuses bitemporal feature interactions, while knowledge distillation may suffer from unreliable teacher supervision under severe domain shifts. Effectively exploiting bitemporal discrepancy cues while preserving historical-domain knowledge therefore remains a key challenge for DICD. A detailed analysis is provided in the Appendix.

\subsection{Overview of the proposed DG-FDD}
To address knowledge confusion and unreliable distillation under domain shifts, we propose DG-FDD, a domain-incremental change detection framework built on a Siamese DeepLabV3+ architecture. As illustrated in Fig. 2, DG-FDD adopts a teacher-student scheme with a frozen teacher and a trainable student. Both networks use an ImageNet-pretrained ResNet-50 encoder with shared weights between the two temporal branches. During incremental learning, the encoder remains frozen, while only the lightweight adaptation modules and the current-domain decoder are updated. To explicitly incorporate bitemporal cues into incremental adaptation, DGDA is introduced into the shallow and intermediate feature extraction stages, where bitemporal feature discrepancies are used to recalibrate adapter responses in a change-sensitive manner. This enables the model to enhance change-relevant channels while suppressing background- and domain-specific interference. The high-level features are further processed by Atrous Spatial Pyramid Pooling (ASPP) and task-specific decoder heads to capture multi-scale context and produce change predictions. To mitigate forgetting under severe domain shifts, DG-FDD further employs a frequency-decoupled dual-track distillation strategy, where cross-domain hybrid features are synthesized by combining historical-domain amplitude priors with current-domain phase information. These synthesized features, together with original features, guide knowledge distillation to improve new-domain adaptation while preserving historical knowledge. By integrating difference-guided adaptation and frequency-domain knowledge transfer, DG-FDD jointly enhances cross-domain change semantic modeling and continual knowledge preservation in DICD.

\begin{figure*}[!h]
\centering

\includegraphics[width=6.4in]{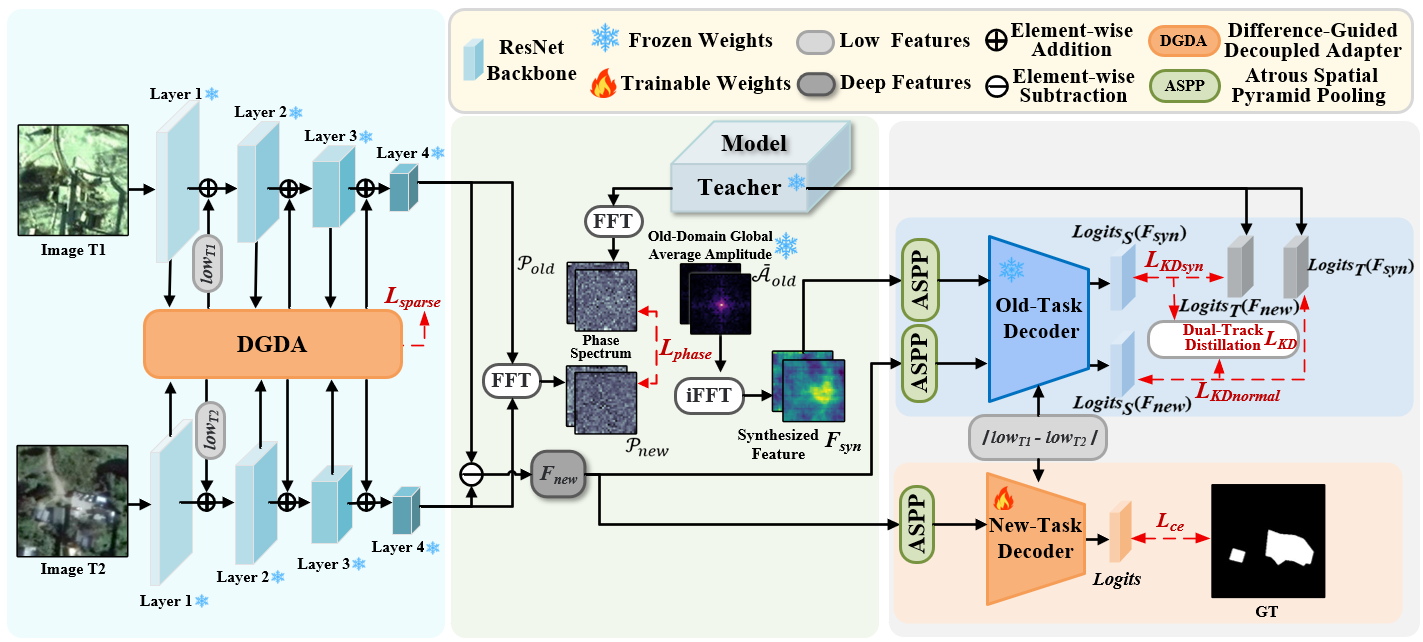}
\caption{Overall framework of the proposed DG-FDD.}
\label{fig:fig2}
\end{figure*}

\subsection{Difference-Guided Dynamic Adapter (DGDA)}
\begin{figure*}[!h]
\centering
\includegraphics[width=6.4in]{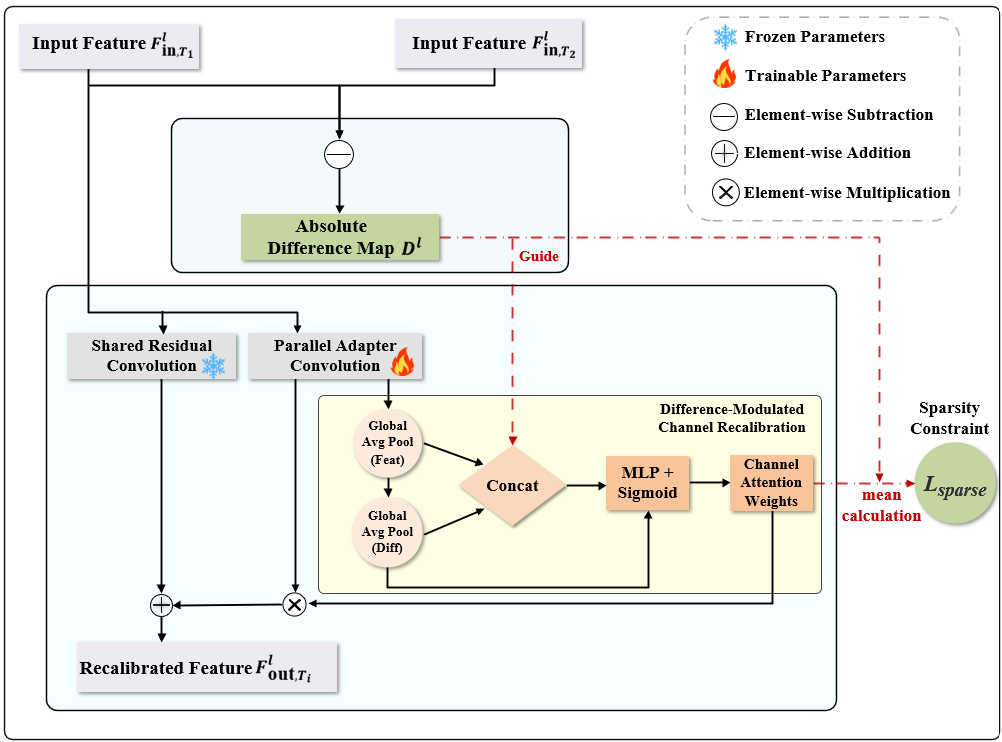}
\caption{Architecture of the DGDA.}
\label{fig:fig3}
\end{figure*}

Traditional parameter-efficient fine-tuning modules designed for single-temporal tasks typically process bitemporal images through separate branches. Under cross-domain incremental scenarios, this temporal independence often causes the adapters to adapt to domain-specific style features rather than robust change semantics. To address this limitation, we propose DGDA. By introducing a bitemporal difference prior, DGDA leverages cross-temporal comparison cues to guide the adapter channels to focus on change-relevant features.
As illustrated in Fig. \ref{fig:fig3}, the DGDA module is embedded in a parallel bypass manner after the residual blocks of the frozen backbone network (Layer~1--Layer~3). Given the feature representations at the $l$-th layer, the DGDA module operates through three main stages: temporal difference perception, joint feature modeling, and difference-guided recalibration. Through these operations, change-relevant information is selectively enhanced, while domain-specific disturbances are suppressed, thereby aiming to enhance the robustness of feature representations under cross-domain incremental learning scenarios.
\subsubsection{Bitemporal difference prior estimation}
Let $F_{\text{in}, T_1}^l, F_{\text{in}, T_2}^l \in \mathbb{R}^{C \times H \times W}$ denote the bitemporal feature maps output by the $l$-th residual block of the frozen backbone. In change detection, genuine change regions typically correspond to significant feature discrepancies between these bitemporal observations, whereas unchanged background regions tend to exhibit consistent semantic representations. To capture these discrepancies, we compute an absolute difference map $D^l$ to serve as a prior indicator of change intensity:
\begin{equation}\label{equ-abs_diff}
	D^l = \left| F_{\text{in}, T_1}^l - F_{\text{in}, T_2}^l \right|
\end{equation}
where $D^l$ encodes the spatial distribution of temporal discrepancies, providing an explicit guidance signal to help suppress domain-specific style variations.

\subsubsection{Joint feature modeling and difference-guided channel attention}
To simultaneously capture individual temporal content and cross-temporal change intensity, we adopt a joint channel-attention modeling strategy. Specifically, global average pooling (GAP) is applied to extract the global contextual descriptor of the $i$-th temporal feature ($i \in \{1, 2\}$) and the global statistical descriptor of the absolute difference map at the $l$-th layer:
\begin{equation}\label{equ-z_feat}
	Z_{\text{feat}, i} = \text{GAP}\left(F_{\text{in}, T_i}^l\right)
\end{equation}
\begin{equation}\label{equ-z_diff}
	Z_{\text{diff}} = \text{GAP}\left(D^l\right)
\end{equation}
where $Z_{\text{feat}, i}$ represents the global contextual descriptor of the $i$-th temporal feature, and $Z_{\text{diff}}$ represents the global statistical discrepancy prior derived from the absolute difference map at the $l$-th layer.

The two descriptors are concatenated along the channel dimension to construct a joint representation that integrates temporal semantic content with the change prior:
\begin{equation}\label{equ-z_joint}
	Z_{\text{joint}, i} = \left[ Z_{\text{feat}, i} \parallel Z_{\text{diff}} \right]
\end{equation}
where $Z_{\text{joint}, i}$ denotes the fused feature vector for the $i$-th temporal branch, and $\parallel$ represents the concatenation operation along the channel dimension. Since both $Z_{\text{feat}, i}$ and $Z_{\text{diff}}$ have a dimensionality of $C$, the joint vector $Z_{\text{joint}, i}$ has a dimensionality of $2C$.

Subsequently, a bottleneck MLP composed of two fully connected layers models the inter-channel dependencies, generating the difference-guided channel attention weight $w_i$ for each temporal branch:
\begin{equation}\label{equ-attn_weight}
	w_i = \sigma\left(W_2 \cdot \delta\left(W_1 \cdot Z_{\text{joint}, i}\right)\right)
\end{equation}
where $\delta(\cdot)$ and $\sigma(\cdot)$ denote the ReLU and Sigmoid activation functions, respectively. $W_1 \in \mathbb{R}^{C/r \times 2C}$ and $W_2 \in \mathbb{R}^{C \times C/r}$ are learnable parameters, and $r$ represents the reduction ratio. This design uses global bitemporal discrepancy statistics to activate change-relevant channels for each branch individually.
\subsubsection{Difference-guided feature recalibration and residual injection}
The channel weight $w_i$ acts as an adaptive gating mechanism. Under the guidance of $Z_{\text{diff}}$, the model is encouraged to assign higher responses to channels associated with pronounced differences while suppressing those with weak variations. The original backbone features are recalibrated using this weight to help filter out change-irrelevant style information:
\begin{equation}\label{equ-recal_feat}
	\tilde{F}_{T_i}^l = F_{\text{in}, T_i}^l \odot w_i
\end{equation}
where $\tilde{F}_{T_i}^l$ represents the recalibrated feature map, and $\odot$ denotes channel-wise multiplication.

Finally, the recalibrated features are injected back into the backbone's data flow via a residual connection, yielding the refined bitemporal feature pair output $\{F_{\text{out}, T_1}^l, F_{\text{out}, T_2}^l\}$:
\begin{equation}\label{equ-out_feat}
	F_{\text{out}, T_i}^l = F_{\text{in}, T_i}^l + \text{Conv}_{1\times 1}\left(\tilde{F}_{T_i}^l\right)
\end{equation}
where $F_{\text{out}, T_i}^l$ represents the refined output feature map for temporal branch $T_i$ ($i \in \{1,2\}$), and $\text{Conv}_{1\times 1}$ denotes a $1 \times 1$ convolutional layer used to align channel dimensions.

\subsubsection{Sparsity constraint strategy}
The channel attention weight $w_{l, i} \in (0,1)^{C_l}$ (for layer $l$ and temporal branch $T_i$) determines the activation level of the adapter parameters. Without explicit regularization, the adapter may activate a large number of channels to overfit fine-grained, domain-specific textures, potentially blending semantic changes with background styles. To address this, we introduce an $L_1$-style sparsity regularization constraint on the attention weights to encourage selective activation.
Specifically, the sparsity loss $\mathcal{L}_{\text{Sparse}}$ is defined as the mean of the attention weights generated across all layers and temporal branches:
\begin{equation}\label{equ-sparsity_loss}
	\mathcal{L}_{\text{Sparse}} = \frac{1}{2L} \sum_{l=1}^{L} \sum_{i=1}^{2} \left( \frac{1}{C_l} \sum_{c=1}^{C_l} w_{l, i, c} \right)
\end{equation}
where $L$ is the number of layers containing DGDA modules, $C_l$ is the channel dimension at layer $l$, and $w_{l, i, c}$ denotes the attention weight of the $c$-th channel of temporal branch $T_i$ at layer $l$.

To prevent the trivial solution where all weights collapse to zero, $\mathcal{L}_{\text{Sparse}}$ is optimized jointly with task-specific losses . This regularization penalizes redundant channel activations, guiding the attention mechanism to selectively activate channels highly correlated with change semantics while suppressing background noise.
\subsection{Frequency-Decoupled Knowledge Distillation with Cross-Domain Feature Synthesis (FDKD-CS)}
Under domain shifts, teacher models trained on historical domains may produce biased representations for new-domain images. Direct knowledge distillation (KD) on such domain-discrepant inputs can introduce unreliable constraints, compromising the stability--plasticity trade-off. To alleviate this supervision degradation, we propose FDKD-CS. Exploiting the frequency characteristics of remote sensing imagery, FDKD-CS disentangles the amplitude spectrum, which mainly encodes domain-specific style variations, from the phase spectrum, which preserves domain-invariant structures and object boundaries. By recombining these components, FDKD-CS generates pseudo-historical style features with new-domain structural information, providing a reliable knowledge transfer pathway without accessing historical data.
To enable cross-domain feature synthesis without violating the data privacy constraint of incremental learning, we introduce pre-stored historical amplitude prototypes, denoted as $\bar{\mathcal{A}}_{\text{old}}^{T_1}$ and $\bar{\mathcal{A}}_{\text{old}}^{T_2}$. Specifically, after training each historical domain, the frozen teacher model is used to extract high-level features from the corresponding domain. The amplitude spectra of the extracted features are obtained via FFT and averaged over all samples to construct the global amplitude prototypes:

\begin{equation}
	\bar{\mathcal{A}}_{\text{old}}^{T_i}
	=
	\frac{1}{N}
	\sum_{n=1}^{N}
	\left|
	\mathcal{F}(F_{n}^{T_i})
	\right|,
	\quad i\in\{1,2\}
\end{equation}

where $F_n^{T_k}$ denotes the high-level feature of the $n$-th historical sample from temporal branch $T_k$, and $N$ is the number of samples used for estimation.

\begin{figure*}[!h]
	\centering
	\includegraphics[width=0.9\linewidth]{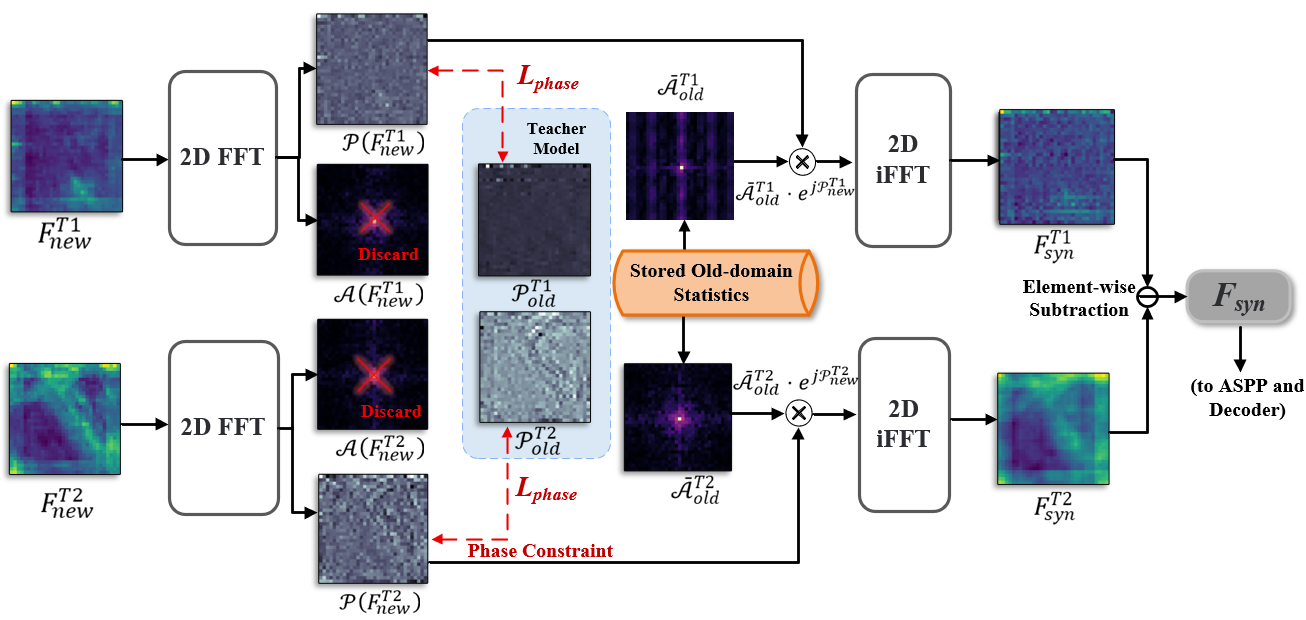}
	\caption{Frequency-domain decoupled cross-domain feature synthesis pipeline.}
	\label{fig:fig4}
\end{figure*}

As illustrated in Fig.~\ref{fig:fig4}, FDKD-CS performs frequency-domain feature synthesis to construct cross-domain representations. Given the new-domain bitemporal features $F_{\text{new}}^{T_1}$ and $F_{\text{new}}^{T_2} \in \mathbb{R}^{C \times H \times W}$ extracted from the student backbone, a two-dimensional Fast Fourier Transform (2D FFT) is first applied to obtain their frequency-domain representations:

\begin{equation}\label{equ-2d_fft}
	\mathcal{F}\left(F_{\text{new}}^{T_i}\right)(u, v) = \sum_{h=0}^{H-1} \sum_{w=0}^{W-1} F_{\text{new}}^{T_i}(h, w) e^{-j 2\pi \left(\frac{uh}{H} + \frac{vw}{W}\right)}
\end{equation}
where $\mathcal{F}\left(F_{\text{new}}^{T_i}\right)(u, v)$ denotes the complex spectral representation of spatial feature map $F_{\text{new}}^{T_i}$ at frequency coordinate $(u,v)$, and $j$ represents the imaginary unit.

These complex representations are further decomposed into amplitude spectra $\mathcal{A}\left(F_{\text{new}}^{T_i}\right)$ and phase spectra $\mathcal{P}\left(F_{\text{new}}^{T_i}\right)$:
\begin{equation}\label{equ-spectrum_decomp}
	\mathcal{A}\left(F_{\text{new}}^{T_i}\right) = \left| \mathcal{F}\left(F_{\text{new}}^{T_i}\right) \right|, \quad \mathcal{P}\left(F_{\text{new}}^{T_i}\right) = \angle \mathcal{F}\left(F_{\text{new}}^{T_i}\right)
\end{equation}
where $|\cdot|$ and $\angle$ denote the modulus operation and the phase extraction operator, respectively.

To inject historical style properties while preserving the spatial structure of the new domain, the style-related amplitude spectra $\mathcal{A}\left(F_{\text{new}}^{T_i}\right)$ are discarded. We then pair the domain-invariant phase spectra $\mathcal{P}\left(F_{\text{new}}^{T_i}\right)$ with the corresponding pre-stored global historical amplitude templates $\bar{\mathcal{A}}_{\text{old}}^{T_i}$. The combined frequency representations are subsequently mapped back to the spatial domain via the 2D inverse Fast Fourier Transform (2D iFFT), yielding the single-temporal synthesized features $F_{\text{syn}}^{T_1}$ and $F_{\text{syn}}^{T_2}$:
\begin{equation}\label{equ-f_syn_single}
	F_{\text{syn}}^{T_i} = \mathcal{F}^{-1}\left( \bar{\mathcal{A}}_{\text{old}}^{T_i} \cdot e^{j \cdot \mathcal{P}\left(F_{\text{new}}^{T_i}\right)} \right)
\end{equation}
where $\mathcal{F}^{-1}(\cdot)$ represents the 2D iFFT, and $e^{j \cdot \mathcal{P}\left(F_{\text{new}}^{T_i}\right)}$ forms the unit-modulus complex representation constructed from the phase spectrum.

After inverse Fourier reconstruction, the synthesized bitemporal features are transformed into difference representations by temporal feature subtraction. Specifically, the synthesized difference feature $F_{\text{syn}}$ and the original new-domain difference feature $F_{\text{new}}$ are obtained as:

\begin{equation}\label{equ-f_fusion}
	F_{\text{syn}} = F_{\text{syn}}^{T_1} - F_{\text{syn}}^{T_2}, \quad
	F_{\text{new}} = F_{\text{new}}^{T_1} - F_{\text{new}}^{T_2}
\end{equation}

The resulting difference representations are subsequently processed by the ASPP module and task-specific decoder heads for semantic prediction.
To regularize the network and mitigate forgetting under severe domain shifts, we apply complementary alignment constraints at both the feature and decision levels. At the feature level, we introduce a phase-based structural consistency loss $\mathcal{L}_{\text{phase}}$ to align the bitemporal spatial structure of the student with the stable structural representations of the teacher:
\begin{equation}\label{equ-phase_loss}
	\mathcal{L}_{\text{phase}} = \frac{1}{2} \left( \left\| \mathcal{P}\left(F_{\text{new}}^{T_1}\right) - \mathcal{P}_{\text{old}}^{T_1} \right\|_2^2 + \left\| \mathcal{P}\left(F_{\text{new}}^{T_2}\right) - \mathcal{P}_{\text{old}}^{T_2} \right\|_2^2 \right)
\end{equation}
where $\mathcal{P}_{\text{old}}^{T_1}$ and $\mathcal{P}_{\text{old}}^{T_2}$ represent the reference phase spectra generated by the teacher model processing the same inputs, and $\|\cdot\|_2^2$ represents the mean squared error. No constraint is imposed on the amplitude spectrum, allowing the student to adaptively fit the style distributions of the new domain.

At the decision level, the dual-track distillation loss $\mathcal{L}_{\text{KD}}$ is enforced on the fused difference features to maintain consistent predictions across both real and synthesized feature distributions:
\begin{equation}\label{equ-kd_loss_normal}
	\mathcal{L}_{\text{KD}_{\text{normal}}} = \mathcal{L}_{\text{KL}}\left(\sigma\left(\text{Logits}_{\text{S}}\left(F_{\text{new}}\right)\right), \sigma\left(\text{Logits}_{\text{T}}\left(F_{\text{new}}\right)\right)\right)
\end{equation}
\begin{equation}\label{equ-kd_loss_syn}
	\mathcal{L}_{\text{KD}_{\text{syn}}} = \mathcal{L}_{\text{KL}}\left(\sigma\left(\text{Logits}_{\text{S}}\left(F_{\text{syn}}\right)\right), \sigma\left(\text{Logits}_{\text{T}}\left(F_{\text{syn}}\right)\right)\right)
\end{equation}
\begin{equation}\label{equ-kd_loss_total}
	\mathcal{L}_{\text{KD}} = \frac{1}{2}\left(\mathcal{L}_{\text{KD}_{\text{normal}}} + \mathcal{L}_{\text{KD}_{\text{syn}}}\right)
\end{equation}
where $\mathcal{L}_{\text{KL}}(\cdot)$ is the Kullback--Leibler divergence, $\sigma(\cdot)$ denotes a normalization operation, and $\text{Logits}_{\text{S}}(\cdot)$ and $\text{Logits}_{\text{T}}(\cdot)$ represent the prediction logits produced by the student and teacher models at the old-task decoder head.

This dual-track formulation ensures consistent decision-level boundaries, effectively promoting the stability-plasticity balance in continual cross-domain change detection.

\subsection{Training Objective and Optimization}
In the DICD framework, learning at stage $t$ requires adapting the model to the current target domain $\mathcal{D}_t$ while preserving discriminative capabilities on previously learned domains $\mathcal{D}_{1:t-1}$ under a data-free constraint. To establish this stability-plasticity balance, we formulate a teacher-student optimization paradigm. The frozen historical model from the previous stage, denoted as $M_{t-1}$ (parameterized by $\theta_{t-1}$), acts as the teacher to provide stable historical supervision. Correspondingly, the current model $M_t$ (parameterized by $\theta_t$, initialized as $\theta_t = \theta_{t-1}$) serves as the student network to be optimized. This dual-model framework guides representation learning by combining task-specific supervised loss with multileveled alignment constraints.
To promote adaptation to the current domain, task-specific supervision is enforced via a pixel-level cross-entropy loss $\mathcal{L}_{\text{CE}}$ on $\mathcal{D}_t$. For each batch of incoming bitemporal image pairs, the student network generates change predictions via the newly appended decoder branch. The task-supervised loss is formulated as:
\begin{equation}\label{equ-ce_loss_task}
	\mathcal{L}_{\text{CE}} = -\frac{1}{N_{\text{pix}}} \sum_{i=1}^{N_{\text{pix}}} \left[ Y_t^{(i)} \log\left(P_t^{(i)}\right) + \left(1 - Y_t^{(i)}\right) \log\left(1 - P_t^{(i)}\right) \right]
\end{equation}
where $Y_t^{(i)} \in \{0,1\}$ denotes the pixel-level ground-truth change label, $P_t^{(i)} = M_t(X^{(i)}; \theta_{\text{new}})$ represents the predicted change probability produced by the student network, and $N_{\text{pix}}$ is the total number of pixels in a training batch.

To preserve historical knowledge during incremental adaptation, three complementary regularization terms are jointly optimized. The feature-level phase consistency loss $\mathcal{L}_{\text{phase}}$ preserves domain-invariant spatial structures under domain shifts, while the dual-track distillation loss $\mathcal{L}_{\text{KD}}$ maintains prediction consistency on both original and synthesized feature distributions. Meanwhile, the sparsity constraint $\mathcal{L}_{\text{Sparse}}$ regularizes adapter activation by promoting selective channel utilization and reducing domain-specific overfitting.
Accordingly, the student network is optimized by jointly considering task adaptation and knowledge preservation. The overall objective at incremental stage $t$ is formulated as:

\begin{equation}\label{equ-total_loss}
	\mathcal{L}_{\text{Total}}
	=
	\mathcal{L}_{\text{CE}}
	+
	\lambda_{\text{ph}}(e)\mathcal{L}_{\text{phase}}
	+
	\mathcal{L}_{\text{KD}}
	+
	\mathcal{L}_{\text{Sparse}},
\end{equation}

where $\lambda_{\text{ph}}(e)$ is an adaptive coefficient controlling the contribution of phase consistency regularization during training.

To avoid restricting early-stage adaptation to the target domain, a step-wise warm-up strategy is adopted for $\lambda_{\text{ph}}(e)$. Specifically, the phase consistency constraint is disabled in the first 20 epochs and activated afterward with a fixed weight of 0.1. The corresponding scheduling function is defined as:

\begin{equation}\label{equ-lambda_scheduler}
	\lambda_{\text{ph}}(e)=
	\begin{cases}
		0, & e\leq20,\\
		0.1, & e>20,
	\end{cases}
\end{equation}

where $e$ denotes the current training epoch. The complete incremental optimization procedure at stage $t$ is summarized in Algorithm~\ref{alg:training_pipeline}.

\begin{algorithm}[!h]
	\caption{Training Procedure of DG-FDD for Domain-Incremental Change Detection}
	\label{alg:training_pipeline}
	\begin{algorithmic}[1]

		\REQUIRE
		Previous model $M_{t-1}$, current-domain data $\mathcal D_t$,
		historical amplitude prototypes $\bar{\mathcal A}_{old}$

		\ENSURE
		Updated model $M_t$

		\STATE Initialize $M_t \leftarrow M_{t-1}$; freeze teacher and backbone

		\FOR{each epoch $e$}
		\FOR{each bitemporal sample $(X_t^{T_1},X_t^{T_2},Y_t)$}

		\STATE Extract hierarchical features with Siamese encoder

		\STATE Generate difference-aware representations using DGDA

		\STATE Construct current-domain difference feature $F_{new}$

		\STATE Perform frequency-decoupled synthesis with
		historical amplitude and current phase to obtain $F_{syn}$

		\STATE Predict change maps using ASPP and task-specific decoder

		\STATE Calculate
		$\mathcal L_{CE}$, $\mathcal L_{KD}$,
		$\mathcal L_{phase}$ and $\mathcal L_{Sparse}$

		\STATE Update $M_t$ by minimizing $\mathcal L_{Total}$

		\ENDFOR
		\ENDFOR

		\STATE Store historical amplitude prototypes for future increments

		\RETURN $M_t$

	\end{algorithmic}
\end{algorithm}
\section{Experimental results and discussion}

\subsection{Datasets description}

Three high-resolution RSCD datasets, i.e., LEVIR-CD~\citep{RN36}, CDD~\citep{RN65}, and GZ-CD~\citep{RN66}, are employed to construct continuous DICD scenarios for evaluating the effectiveness of the proposed method. The characteristics and corresponding data splits of these datasets are summarized in Table 1. As illustrated in Fig. 5, these datasets exhibit considerable variations in land-cover distributions, background textures, and imaging conditions, reflecting the inherent heterogeneity across different remote sensing scenarios.
To further quantify the domain discrepancy, Figs. 6--7 analyze the low-level visual statistics and high-level semantic feature distributions, respectively. The results demonstrate evident shifts in both color distributions and deep feature embeddings, confirming substantial cross-domain discrepancies among different datasets. Accordingly, the three datasets are sequentially regarded as incremental domains $\mathcal{D}_1$, $\mathcal{D}_2$, and $\mathcal{D}_3$ to simulate realistic continuous DICD.
\par
\begin{table}[!h]
	\centering
	\small
	\setlength{\tabcolsep}{4.5pt}
	\caption{Detailed characteristics and split statistics of the three experimental datasets.}
	\label{tab:datasets}
	\begin{tabular}{lccc}
		\toprule
		\textbf{Characteristic} & \textbf{LEVIR-CD} & \textbf{CDD} & \textbf{GZ-CD} \\
		\midrule
		Image Mode & RGB & RGB & RGB \\
		Capturing Time & 2002--2018 & Seasonal variations & 2006--2019 \\
		Included Location & Texas, USA & Google Earth (Global) & Guangzhou, China \\
		Spatial Resolution & 0.5 m/pixel & 0.03--1.0 m/pixel & 0.55 m/pixel \\
		Original Size & $1024 \times 1024$ & \begin{tabular}[c]{@{}c@{}}$4725 \times 2700$, \\ $1900 \times 1000$\end{tabular} & \begin{tabular}[c]{@{}c@{}}$1006 \times 1168$ to \\ $4936 \times 5224$\end{tabular} \\
		Original Pairs & 637 & 11 & 19 \\
		Cropped Size & $256 \times 256$ & $256 \times 256$ & $256 \times 256$ \\
		Total Cropped Pairs & 10,192 & 16,000 & 3,443 \\
		\midrule
		Dataset Split Pairs & 7,120 : 1,024 : 2,048 & 10,000 : 3,000 : 3,000 & 2,753 : 345 : 345 \\
		(Train : Val : Test) & ($\approx 70\% : 10\% : 20\%$) & ($62.5\% : 18.75\% : 18.75\%$) & ($\approx 80\% : 10\% : 10\%$) \\
		\bottomrule
	\end{tabular}
\end{table}

\begin{figure*}[!h]
\centering
\includegraphics[width=0.8\linewidth]{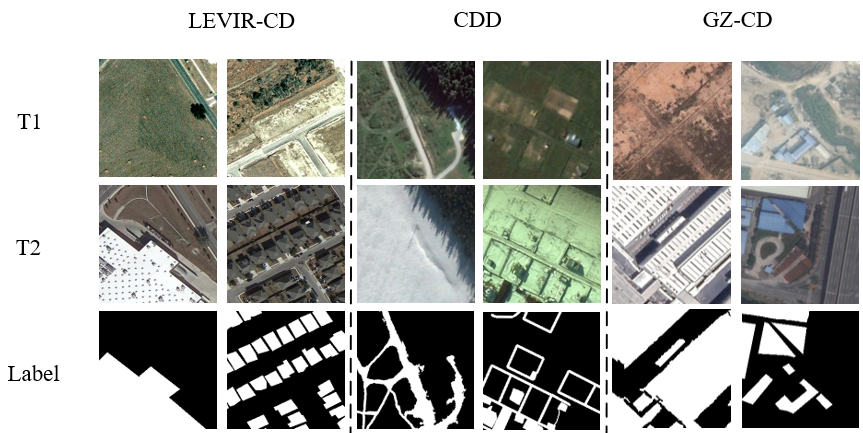}
\caption{Multi-domain incremental change detection datasets.}
\label{fig:fig5}
\end{figure*}

\begin{figure*}[!h]
\centering
\includegraphics[width=\linewidth]{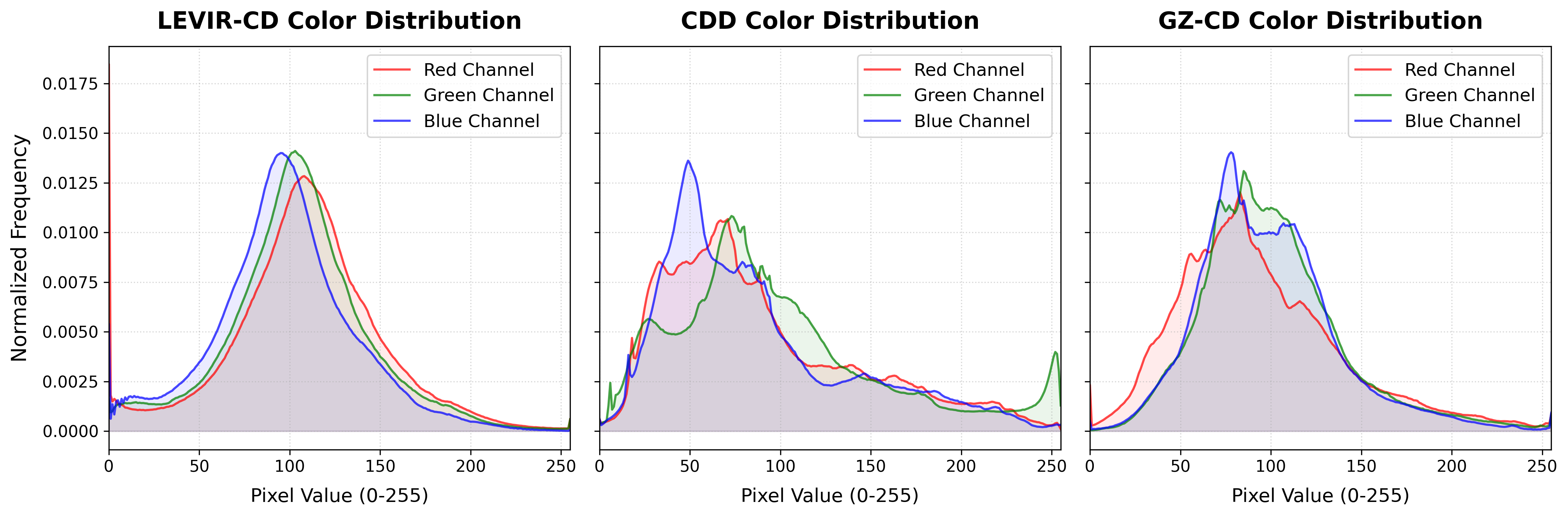}
\caption{Comparison of normalized color frequency in RGB channels among LEVIR-CD, CDD, and GZ-CD datasets.}
\label{fig:fig6}
\end{figure*}

\begin{figure*}[!h]
	\centering
	\includegraphics[width=0.6\linewidth]{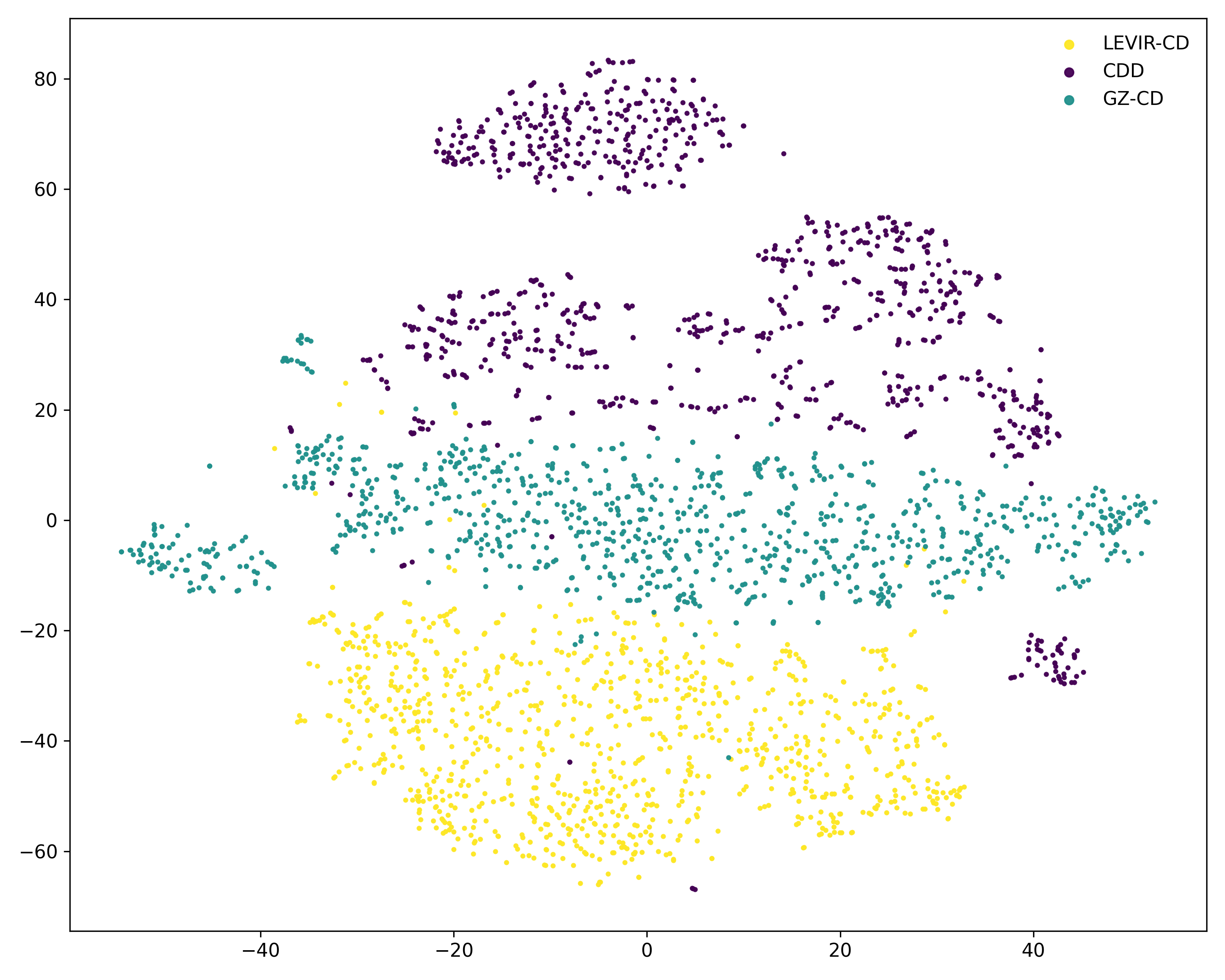}
	\caption{t-distributed stochastic neighbor embedding (t-SNE) visualization of the high-level semantic feature distributions across LEVIR-CD, CDD, and GZ-CD.}
	\label{fig:sne}
\end{figure*}

\subsection{Implementation details}
The proposed DG-FDD framework and all baseline methods are implemented in PyTorch. Experiments are conducted on two NVIDIA GeForce RTX 4060~Ti GPUs with 16~GB memory each. Both the student and teacher networks use ImageNet-pretrained ResNet-50 backbones, where the residual blocks and batch normalization layers are frozen throughout continual learning to reduce parameter-update interference. Only the DGDA modules and task-specific decoder heads are optimized. Random horizontal flipping, vertical flipping, and rotation are applied for data augmentation. To simulate real-world streaming scenarios, two-task and three-task cross-domain incremental settings are constructed, where models are trained sequentially across domains (e.g., LEVIR-CD $\rightarrow$ CDD $\rightarrow$ GZ-CD) without accessing raw data from previous domains. For each task, the model is trained for 150~epochs with a batch size of 8 using the AdamW optimizer, with an initial learning rate of $5 \times 10^{-4}$ and a weight decay of $1 \times 10^{-4}$. A LambdaLR scheduler is used for smooth convergence, and the frequency-domain phase consistency loss is activated after a 20-epoch warm-up with its weight fixed at 0.1.

\subsection{Evaluation metrics}

For each independent incremental task, four widely used metrics in CD are adopted for quantitative evaluation, including Precision, Recall, Intersection over Union (IoU), and F1-score. The corresponding formulations are given as follows:
\begin{equation}
	\text{Precision} = \frac{TP}{TP + FP}
\end{equation}
\begin{equation}
	\text{Recall} = \frac{TP}{TP + FN}
\end{equation}
\begin{equation}
	F1 = \frac{2}{\text{recall}^{-1} + \text{precision}^{-1}}
\end{equation}
\begin{equation}
	\text{IoU} = \frac{TP}{TP + FN + FP}
\end{equation}
where, True Positive ($TP$) denotes correctly classified changed pixels, False Positive ($FP$) represents unchanged pixels incorrectly predicted as changed, True Negative ($TN$) indicates correctly classified unchanged pixels, and False Negative ($FN$) refers to missed changed pixels.

While these conventional metrics evaluate the performance of a model on a specific domain, they fail to capture the degree of catastrophic forgetting in incremental learning. To quantify knowledge retention on previous domains after  learning new ones, we follow the evaluation protocol of~\citep{RN27} and introduce the average performance degradation metrics $\Delta\text{F1}$ and $\Delta\text{IoU}$, which are defined as:
\begin{equation}
	\Delta\text{IoU} = \frac{1}{T} \sum_{t=1}^{T} \frac{\text{IoU}_{m,t} - \text{IoU}_{b,t}}{\text{IoU}_{b,t}}
\end{equation}
\begin{equation}
	\Delta\text{F1} = \frac{1}{T} \sum_{t=1}^{T} \frac{\text{F1}_{m,t} - \text{F1}_{b,t}}{\text{F1}_{b,t}}
\end{equation}
where $T$ denotes the total number of learned tasks. $\text{IoU}_{m,t}$ and $\text{F1}_{m,t}$ represent the performance of the incremental model $m$ on task $t$ after completing all training stages, while $\text{IoU}_{b,t}$ and $\text{F1}_{b,t}$ denote the corresponding results of the single-task baseline model $b$.

A smaller $\Delta\text{F1}$ or $\Delta\text{IoU}$ indicates that the incremental model is closer to the upper-bound performance achieved by single-task training, suggesting milder forgetting and a better stability--plasticity trade-off. In contrast, larger values indicate more severe degradation of previously learned knowledge.

\subsection{Competitors}

To evaluate the effectiveness of DG-FDD, we compare it with representative CD and IL methods, including conventional training strategies, CD networks, and DIL approaches.

\begin{itemize}
	\item[1)] Single-task: Models are independently trained and evaluated on each domain using the full training data. Without incremental learning, this setting is unaffected by catastrophic forgetting and serves as an empirical upper bound for single-domain performance.

	\item[2)] FT: A basic IL baseline that sequentially fine-tunes the model on each new domain without any forgetting-mitigation mechanism.

	\item[3)] FE: A parameter-freezing baseline that fixes the feature extractor after source-domain training and updates only the task-specific decoder during incremental learning.

	\item[4)] BIT~\citep{RN41}: A Transformer-based CD method that models global contextual relationships between bitemporal features for change detection.

	\item[5)] STRobustNet~\citep{RN67}: A robust CD method that addresses bitemporal spatiotemporal inconsistencies through global and sample-level contextual modeling.

	\item[6)] MDIL~\citep{RN26}: A DIL method that dynamically isolates domain-invariant and domain-specific parameters to balance knowledge retention and adaptation.

	\item[7)] MDINet~\citep{RN27}: A DIL method for multidomain CD that mitigates cross-domain knowledge interference through domain residual adaptation and hierarchical knowledge distillation.

	\item[8)] FOCUS~\citep{RN68}: A feature-replay-based IL method that stores historical features and imposes feature consistency constraints to mitigate catastrophic forgetting.

	\item[9)] DI-FRCNN~\citep{RN60}: A dual-teacher DIL method that combines pseudo-label supervision and knowledge-transfer constraints to facilitate new-domain adaptation while preserving previously learned knowledge.
\end{itemize}

\subsection{Comparative analysis}
\subsubsection{Two-domain incremental learning analysis}
Table~2 summarizes the results on the LEVIR $\rightarrow$ CDD two-domain incremental sequence, where LEVIR and CDD denote the historical and new domains after IL, respectively. The Single-task model provides the upper-bound performance by independently optimizing each domain.
Direct fine-tuning (FT) achieves strong adaptation to CDD but incurs severe forgetting on LEVIR, reducing the F1 score to only $12.55\%$. Similar degradation is observed in BIT and STRobustNet, demonstrating that feature adaptation alone cannot effectively preserve historical knowledge under domain shifts. The feature extraction strategy (FE) alleviates forgetting by freezing most parameters and retains $89.16\%$ F1 on LEVIR, but its limited parameter updates restrict adaptation to CDD. Continual learning methods, including MDIL, FOCUS, and DI-FRCNN, provide improved stability-plasticity balance, yet still exhibit considerable performance degradation, with DI-FRCNN, MDIL, and FOCUS suffering decreases of $-29.70\%$ in $\Delta\text{F1}$, $-31.12\%$ in $\Delta\text{IoU}$, and $-23.25\%$ in $\Delta\text{IoU}$, respectively. MDINet further reduces forgetting through knowledge distillation, achieving relatively competitive performance with $\Delta\text{F1}$ and $\Delta\text{IoU}$ of $-4.01\%$ and $-7.03\%$.
In comparison, the proposed method achieves a more favorable stability-plasticity balance. After IL, it obtains F1 scores of $90.40\%$ and $96.75\%$ on LEVIR and CDD, respectively, with only $0.05\%$ and $1.08\%$ decreases from the corresponding Single-task models. The average degradation is limited to $\Delta\text{F1}=-0.58\%$ and $\Delta\text{IoU}=-1.12\%$, demonstrating superior knowledge retention while preserving effective adaptation to the new domain.
\begin{table}[htbp]
	\centering
	\caption{Detailed results of incremental experiments conducted on the LEVIR $\rightarrow$ CDD sequence for the proposed method and competing approaches. The arrow indicates the learning order.}
	\label{tab:levir_cdd_results}
	\begin{tabular}{lcccc}
		\toprule
		\textbf{Methods} & \textbf{F1 (\%)} & \textbf{$\Delta$F1 (\%)} & \textbf{IoU (\%)} & \textbf{$\Delta$IoU (\%)} \\
		& LEVIR $\rightarrow$ CDD & & LEVIR $\rightarrow$ CDD & \\
		\midrule
		Single-task & $\mathbf{90.45 \phantom{\rightarrow} 97.83}$ & -- & $\mathbf{82.57 \phantom{\rightarrow} 95.75}$ & -- \\
		\midrule
		BIT & $40.69 \rightarrow 64.34$ & $-44.62$ & $25.55 \rightarrow 47.43$ & $-59.76$ \\
		STRobustNet & $23.77 \rightarrow 89.67$ & $-41.03$ & $13.49 \rightarrow 81.27$ & $-49.39$ \\
		FT & $12.55 \rightarrow 97.02$ & $-43.48$ & $6.69 \rightarrow 94.20$ & $-46.76$ \\
		FE & $89.16 \rightarrow 78.40$ & $-10.64$ & $80.44 \rightarrow 64.47$ & $-17.77$ \\
		MDIL & $72.18 \rightarrow 79.83$ & $-19.30$ & $56.47 \rightarrow 66.42$ & $-31.12$ \\
		MDINet & $84.14 \rightarrow 96.81$ & $-4.01$ & $72.63 \rightarrow 93.82$ & $-7.03$ \\
		FOCUS & $73.96 \rightarrow 88.22$ & $-14.03$ & $58.68 \rightarrow 78.93$ & $-23.25$ \\
		DI-FRCNN & $53.61 \rightarrow 79.57$ & $-29.70$ & $36.62 \rightarrow 66.07$ & $-43.32$ \\
		\midrule
		\textbf{DG-FDD} & $90.40 \rightarrow 96.75$ & $\mathbf{-0.58}$ & $82.48 \rightarrow 93.71$ & $\mathbf{-1.12}$ \\
		\bottomrule
	\end{tabular}
\end{table}
\begin{figure}[!h]
	\centering
	\includegraphics[width=\textwidth]{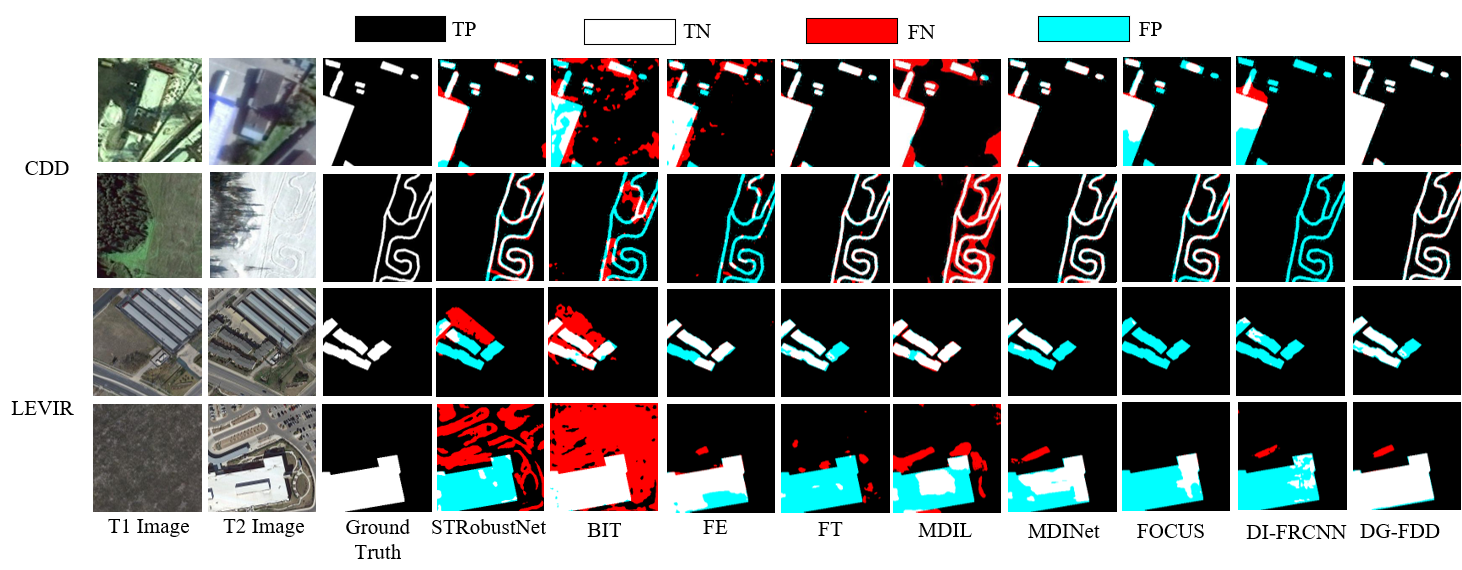}
	\caption{Visual comparison of the proposed method and competing methods on the LEVIR $\rightarrow$ CDD incremental sequence.}
	\label{fig:fig8}
\end{figure}
Fig.~8 presents qualitative comparisons under the LEVIR $\rightarrow$ CDD IL setting, where cyan and red regions denote omission and commission errors, respectively. In the new-domain CDD scenario, most existing methods exhibit degraded predictions under significant domain shifts caused by seasonal changes, snow coverage, and complex background textures. These variations mainly introduce commission errors in snow-covered and shadowed areas, as well as omission errors for small-scale building changes. For example, BIT and MDIL tend to generate false alarms in snow-covered regions, whereas FE, FOCUS, and DI-FRCNN produce fragmented predictions and inaccurate boundaries around narrow roads and building footprints, revealing limited adaptability to substantial appearance variations. In contrast, the proposed method effectively suppresses background-induced disturbances and achieves more complete change regions with refined spatial boundaries.
In the historical-domain LEVIR scenario, several methods suffer from noticeable knowledge degradation after incremental learning. BIT and STRobustNet show reduced sensitivity to previously learned change patterns, resulting in missing detections of certain building changes. Although MDIL, MDINet, and DI-FRCNN alleviate forgetting to some extent, residual omission errors and false responses remain around large building boundaries and shadow regions, indicating incomplete retention of historical representations. By comparison, the proposed method preserves consistent detection capability across both domains, with improved structural completeness and boundary continuity, demonstrating its effectiveness in mitigating catastrophic forgetting while adapting to new-domain characteristics.

To further assess the cross-domain generalization capability of the proposed method, Table~3 summarizes the results on six two-domain incremental learning sequences. Across all scenarios, competing methods exhibit different levels of performance degradation, whereas DG-FDD maintains stable performance under diverse domain transitions. Except for the GZ $\rightarrow$ CDD sequence, the performance variations of DG-FDD remain within $-1.5\%$ in terms of both $\Delta\text{F1}$ and $\Delta\text{IoU}$, demonstrating its effectiveness in preserving historical knowledge while adapting to unseen domains.
Furthermore, positive transfer is observed in the LEVIR $\rightarrow$ GZ and CDD $\rightarrow$ GZ sequences, where the incremental model slightly exceeds the corresponding Single-task upper bounds. This suggests that knowledge learned from previous domains can provide beneficial representations for subsequent adaptation, rather than merely preventing catastrophic forgetting. Overall, these results verify the robustness of DG-FDD under heterogeneous domain shifts and highlight the advantage of jointly optimizing domain adaptation and knowledge preservation in cross-domain continual learning.
\begin{table*}[htbp]
	\centering
	\caption{Quantitative comparison under six two-domain cross-domain incremental sequences.}
	\label{tab:two_task_results}
	\small
	\begin{tabular*}{\textwidth}{@{\extracolsep{\fill}}lccc}
		\toprule
		\textbf{Methods} & \textbf{LEVIR$\rightarrow$CDD} & \textbf{CDD$\rightarrow$LEVIR} & \textbf{GZ$\rightarrow$LEVIR} \\
		& $\Delta$F1 (\%) / $\Delta$IoU (\%) & $\Delta$F1 (\%) / $\Delta$IoU (\%) & $\Delta$F1 (\%) / $\Delta$IoU (\%) \\
		\midrule
		BIT & $-44.62$ / $-59.76$ & $-49.10$ / $-60.00$ & $-40.82$ / $-54.93$ \\
		STRobustNet & $-41.03$ / $-46.76$ & $-38.79$ / $-44.44$ & $-33.32$ / $-39.88$ \\
		FT & $-43.48$ / $-49.39$ & $-39.25$ / $-44.76$ & $-44.94$ / $-50.28$ \\
		FE & $-10.64$ / $-17.77$ & $-5.24$ / $-8.94$ & $-9.50$ / $-15.86$ \\
		MDIL & $-19.30$ / $-31.12$ & $-18.66$ / $-28.47$ & $-30.35$ / $-41.93$ \\
		MDINet & $-4.01$ / $-7.03$ & $-1.75$ / $-3.27$ & $-8.29$ / $-13.10$ \\
		FOCUS & $-14.03$ / $-23.25$ & $-22.56$ / $-34.13$ & $-42.33$ / $-50.39$ \\
		DI-FRCNN & $-29.70$ / $-43.32$ & $-19.10$ / $-30.85$ & $-50.91$ / $-64.66$ \\
		\midrule
		\textbf{DG-FDD} & $\mathbf{-0.58}$ / $\mathbf{-1.12}$ & $\mathbf{-0.52}$ / $\mathbf{-1.00}$ & $\mathbf{-0.67}$ / $\mathbf{-1.18}$ \\
		\bottomrule
	\end{tabular*}

	\vspace{1.5em}

	\begin{tabular*}{\textwidth}{@{\extracolsep{\fill}}lccc}
		\toprule
		\textbf{Methods} & \textbf{LEVIR$\rightarrow$GZ} & \textbf{GZ$\rightarrow$CDD} & \textbf{CDD$\rightarrow$GZ} \\
		& $\Delta$F1 (\%) / $\Delta$IoU (\%) & $\Delta$F1 (\%) / $\Delta$IoU (\%) & $\Delta$F1 (\%) / $\Delta$IoU (\%) \\
		\midrule
		BIT & $-18.14$ / $-28.40$ & $-60.78$ / $-73.17$ & $-36.62$ / $-48.78$ \\
		STRobustNet & $-51.51$ / $-58.14$ & $-48.79$ / $-57.34$ & $-58.97$ / $-69.72$ \\
		FT & $-55.33$ / $-59.17$ & $-45.70$ / $-51.06$ & $-40.18$ / $-47.59$ \\
		FE & $-5.43$ / $-9.19$ & $-10.14$ / $-17.31$ & $-3.96$ / $-16.70$ \\
		MDIL & $-16.28$ / $-25.56$ & $-27.08$ / $-40.77$ & $-41.57$ / $-50.07$ \\
		MDINet & $-0.58$ / $-1.02$ & $-16.22$ / $-23.53$ & $-0.86$ / $-1.65$ \\
		FOCUS & $-13.10$ / $-21.20$ & $-36.02$ / $-45.37$ & $-11.49$ / $-19.52$ \\
		DI-FRCNN & $-29.49$ / $-30.48$ & $-21.59$ / $-33.39$ & $-11.55$ / $-19.31$ \\
		\midrule
		\textbf{DG-FDD} & $\mathbf{0.82}$ / $\mathbf{1.46}$ & $\mathbf{-1.10}$ / $\mathbf{-2.02}$ & $\mathbf{0.67}$ / $\mathbf{1.16}$ \\
		\bottomrule
	\end{tabular*}
\end{table*}
\begin{figure}[htbp]
	\centering

	\includegraphics[width=\textwidth]{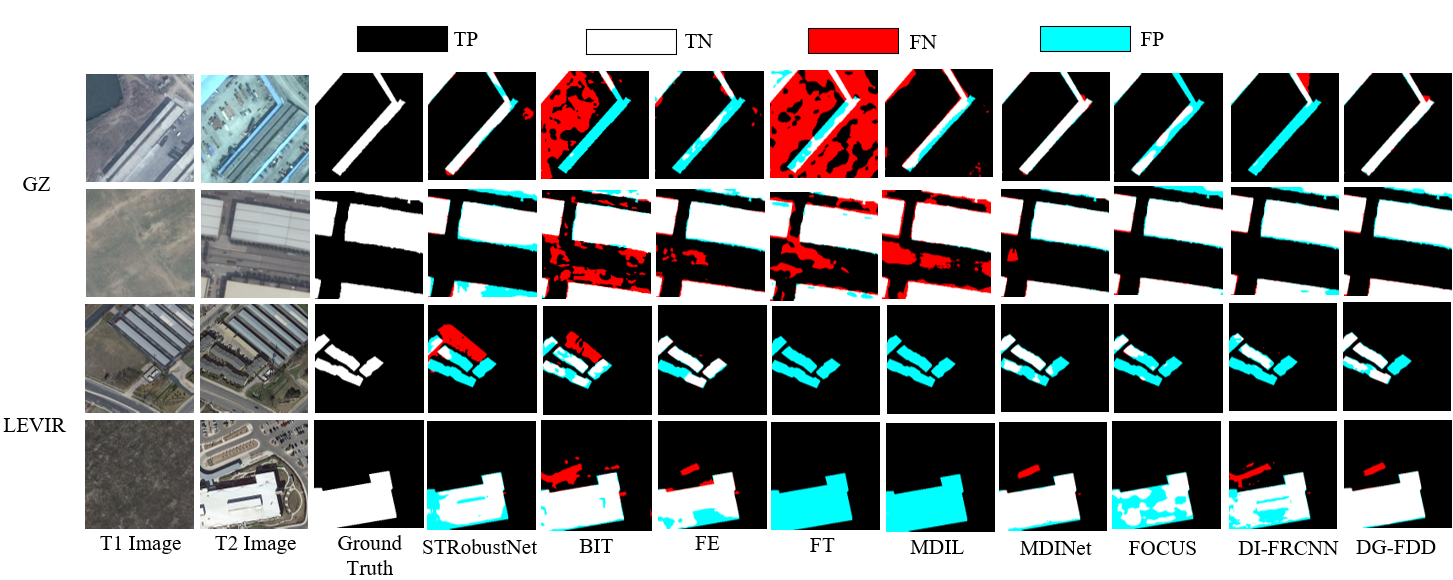}
	\caption{Visual comparison of the proposed method and competing methods on the LEVIR $\rightarrow$ GZ incremental sequence.}
	\label{fig:fig9}
\end{figure}
Fig.~9 presents qualitative comparisons under the LEVIR $\rightarrow$ GZ IL setting. Compared with the LEVIR $\rightarrow$ CDD scenario, the domain discrepancy between LEVIR and GZ is relatively moderate, providing an additional evaluation of model transferability under reduced domain shifts.
In the new-domain GZ scenario, several methods still exhibit false detections caused by variations in urban background appearance and spatial textures. FT and BIT generate noticeable commission errors in complex background regions, while STRobustNet, FE, and MDIL produce more stable predictions but suffer from localized boundary deviations around building footprints. Although MDINet, FOCUS, and DI-FRCNN further alleviate these issues, fragmented structures and minor discontinuities remain in challenging areas. In comparison, the proposed method achieves more consistent change localization with improved structural continuity and boundary alignment, effectively reducing both false alarms and missed detections.
Together with the LEVIR $\rightarrow$ CDD results, these observations demonstrate that DG-FDD maintains reliable adaptation capability across different levels of domain discrepancy, validating its robustness under diverse cross-domain IL scenarios.
\subsubsection{Three-domain continual cross-domain incremental analysis}
Table~4 summarizes the results under the more challenging three-domain IL setting. Compared with the two-domain scenario, sequential adaptation across multiple domains introduces accumulated distribution shifts, leading to increased feature drift and more severe forgetting in most competing methods.
As shown in Table~4, conventional approaches such as BIT and STRobustNet suffer from severe performance degradation, with $\Delta\text{IoU}$ dropping by more than $50\%$ in multiple sequences. Other methods, including MDIL, FOCUS, and DI-FRCNN, also exhibit consistent performance deterioration as domain shifts accumulate. Although MDINet achieves relatively better stability, its $\Delta\text{IoU}$ still degrades to $-7.03\%$, $-1.61\%$, and $-2.45\%$ across different sequences. In contrast, the proposed method maintains consistent performance across all three-domain settings, achieving $\Delta\text{F1}$ values of $-0.49\%$, $-0.26\%$, and $-1.32\%$, and $\Delta\text{IoU}$ values of $-0.99\%$, $-0.54\%$, and $-2.40\%$, respectively. These results demonstrate that DG-FDD effectively mitigates cumulative forgetting during long-term domain evolution.
\begin{table}[htbp]
	\centering
	\caption{Quantitative comparison under three-domain continual cross-domain incremental sequences.}
	\label{tab:three_task_results}
	\begin{tabular}{lccc}
		\toprule
		\textbf{Methods} & \textbf{LEVIR$\rightarrow$CDD$\rightarrow$GZ} & \textbf{LEVIR$\rightarrow$GZ$\rightarrow$CDD} & \textbf{CDD$\rightarrow$LEVIR$\rightarrow$GZ} \\
		& $\Delta$F1 (\%) \& $\Delta$IoU (\%) & $\Delta$F1 (\%) \& $\Delta$IoU (\%) & $\Delta$F1 (\%) \& $\Delta$IoU (\%) \\
		\midrule
		BIT & $-40.86$ \& $-53.00$ & $-49.82$ \& $-64.22$ & $-50.03$ \& $-60.97$ \\
		STRobustNet & $-55.25$ \& $-62.84$ & $-63.32$ \& $-69.82$ & $-55.26$ \& $-62.94$ \\
		FT & $-43.48$ \& $-46.76$ & $-67.04$ \& $-65.73$ & $-54.91$ \& $-60.63$ \\
		FE & $-10.64$ \& $-17.77$ & $-10.21$ \& $-16.97$ & $-6.70$ \& $-11.24$ \\
		MDIL & $-19.30$ \& $-31.12$ & $-51.46$ \& $-63.33$ & $-35.20$ \& $-47.50$ \\
		MDINet & $-4.01$ \& $-7.03$ & $-0.88$ \& $-1.61$ & $-1.43$ \& $-2.45$ \\
		FOCUS & $-27.71$ \& $-37.19$ & $-22.48$ \& $-33.72$ & $-22.15$ \& $-31.42$ \\
		DI-FRCNN & $-20.98$ \& $-32.45$ & $-34.73$ \& $-47.84$ & $-27.39$ \& $-39.61$ \\
		\midrule
		\textbf{DG-FDD} & \textbf{$\mathbf{-0.49}$ \& $\mathbf{-0.99}$} & \textbf{$\mathbf{-0.26}$ \& $\mathbf{-0.54}$} & \textbf{$\mathbf{-1.32}$ \& $\mathbf{-2.40}$} \\
		\bottomrule
	\end{tabular}
\end{table}
\begin{figure}[!h]
	\centering

	\includegraphics[width=\textwidth]{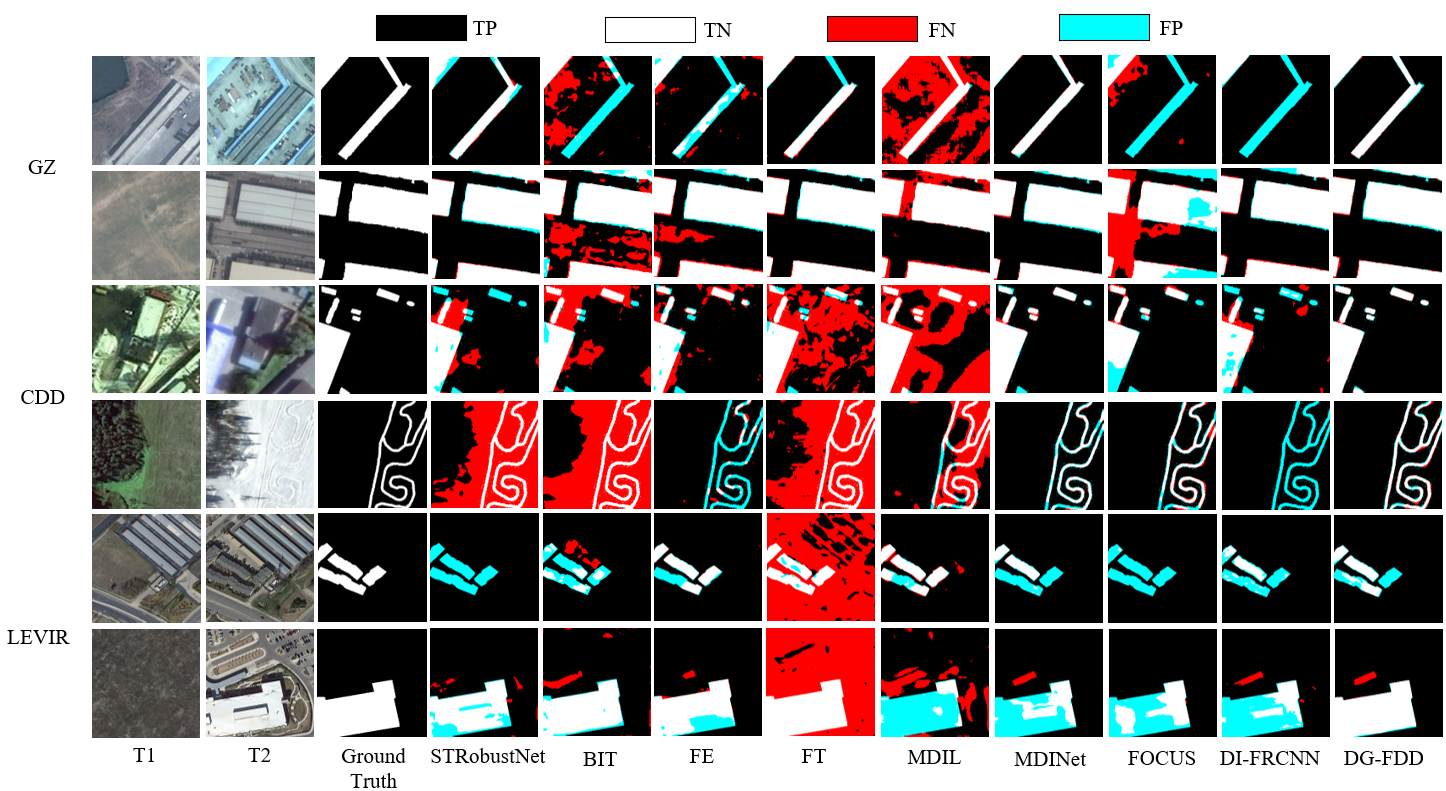}
	\caption{Visual comparison of the proposed method and competing methods on the LEVIR $\rightarrow$ CDD $\rightarrow$ GZ incremental sequence.}
	\label{fig:fig10}
\end{figure}
Fig.~10 presents qualitative comparisons under the three-domain IL setting. With the accumulation of domain shifts, most competing methods exhibit progressive degradation in prediction quality, characterized by fragmented change regions, incomplete object structures, and inconsistent boundaries. These visual artifacts indicate the difficulty of preserving stable change representations during sequential domain adaptation.
In contrast, the proposed method maintains more consistent change localization throughout all incremental stages, with improved object completeness and boundary continuity. It effectively preserves structural details in complex background areas and large-scale building regions while reducing both false alarms and missed detections. These qualitative results are consistent with the quantitative evaluations in Table~4, further demonstrating the capability of DG-FDD to alleviate cumulative forgetting in long-term cross-domain continual learning.

\subsection{Discussion}
\subsubsection{Effect of the phase-consistency loss weight under domain shifts}
To investigate the impact of the phase-consistency loss weight $\lambda_{ph}$ on IL performance, a sensitivity analysis is conducted on the LEVIR $\rightarrow$ GZ sequence. Fig.~11 illustrates the variations of the average performance degradation metrics, $\Delta\text{F1}$ and $\Delta\text{IoU}$, under different $\lambda_{ph}$ configurations.
The results reveal a nonlinear relationship between $\lambda_{ph}$ and incremental performance. With a small $\lambda_{ph}$, the phase-consistency constraint provides insufficient structural regularization, limiting the preservation of historical representations and resulting in noticeable performance degradation. As $\lambda_{ph}$ increases to $0.1$, both $\Delta\text{F1}$ and $\Delta\text{IoU}$ achieve the best performance and become positive, indicating potential positive transfer by effectively preserving shared structural characteristics while allowing sufficient adaptation to the new domain.
However, excessively increasing $\lambda_{ph}$ introduces overly strong structural constraints. When $\lambda_{ph}=1$, the $\Delta\text{IoU}$ decreases by more than $30\%$, suggesting that excessive emphasis on structural consistency impedes adaptation to new-domain distributions and reduces model flexibility. Therefore, $\lambda_{ph}=0.1$ provides an appropriate balance between historical knowledge preservation and new-domain adaptation, confirming the effectiveness of the proposed phase-consistency constraint in cross-domain IL.
\begin{figure}[htbp]
	\centering

	\includegraphics[width=0.7\textwidth]{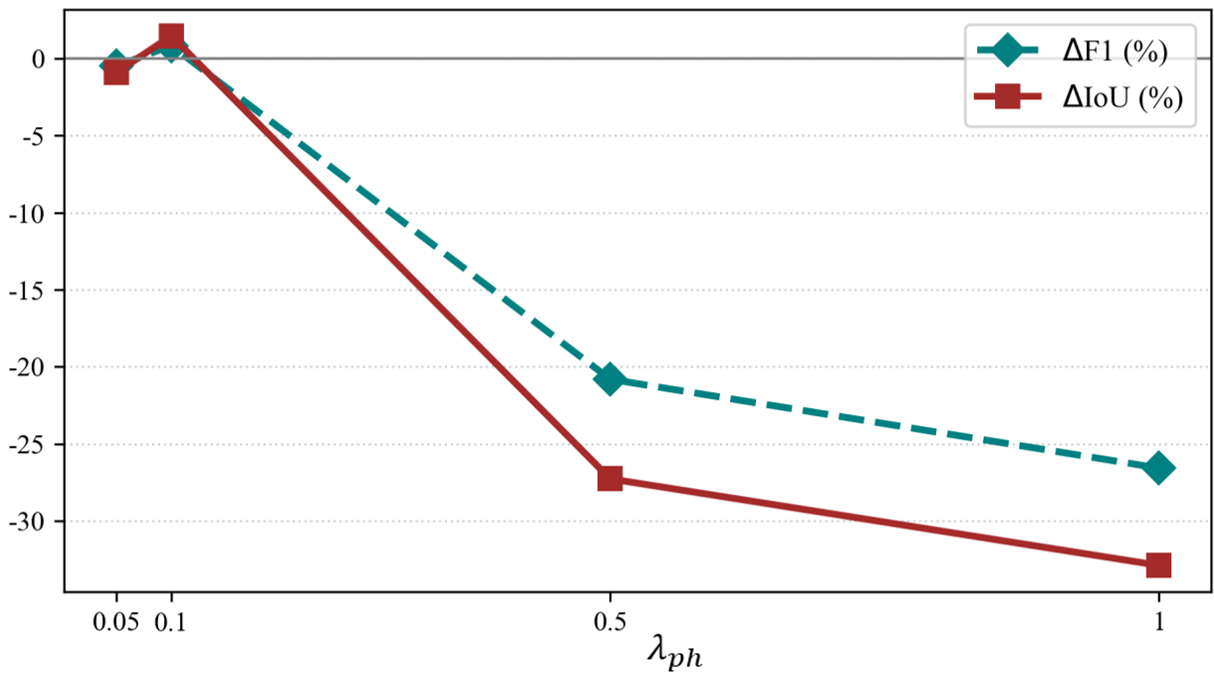}
	\caption{Variations of $\Delta\text{F1}$ and $\Delta\text{IoU}$ with respect to $\lambda_{ph}$ on the LEVIR $\rightarrow$ GZ sequence.}
	\label{fig:fig11}
\end{figure}
\subsubsection{Effect of the frequency-decoupled distillation strategy under domain shifts}
This section investigates the influence of different KD strategies on incremental learning performance under domain shifts. Existing approaches generally rely on either response-based distillation (KD-FC) or intermediate feature-based distillation (KD-HD) to transfer historical knowledge. Fig.~12 compares the average performance degradation metrics, $\Delta\text{F1}$ and $\Delta\text{IoU}$, under different distillation schemes, where zero indicates no degradation relative to the Single-task reference.
As shown in Fig.~12, KD-FC exhibits considerable performance degradation under significant domain shifts, mainly because biased teacher predictions on new-domain samples provide unreliable supervision signals. KD-HD improves knowledge retention by introducing feature-level constraints; however, the transferred representations remain affected by cross-domain distribution discrepancies. In contrast, FDKD-CS achieves positive gains of $0.82\%$ in $\Delta\text{F1}$ and $1.46\%$ in $\Delta\text{IoU}$ on the LEVIR $\rightarrow$ GZ sequence. By decoupling domain-specific frequency components and synthesizing cross-domain features, FDKD-CS reduces distribution mismatch during knowledge transfer and establishes a more favorable stability-plasticity balance without requiring historical data replay.
\begin{figure}[htbp]
	\centering

	\includegraphics[width=0.7\textwidth]{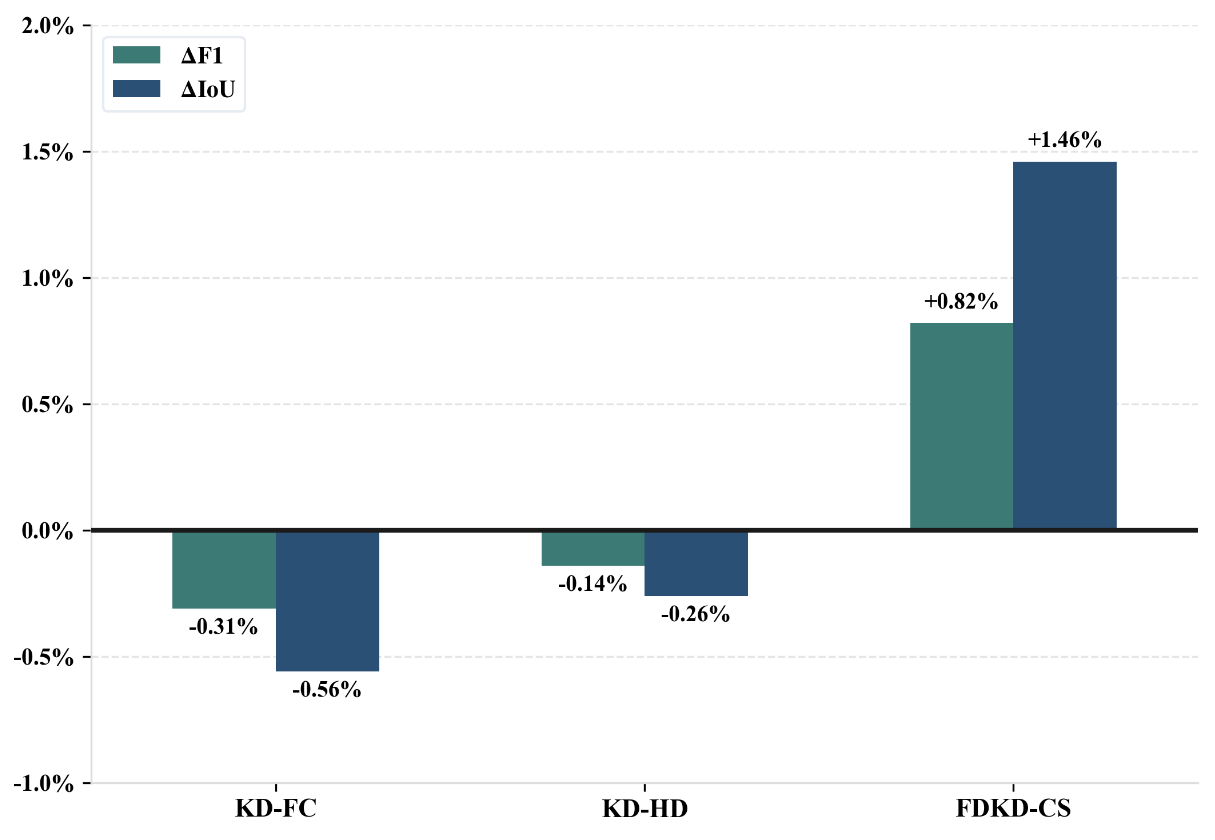}
	\caption{Comparison of performance change rates $\Delta\text{F1}$ and $\Delta\text{IoU}$ under different knowledge distillation strategies on the LEVIR $\rightarrow$ GZ sequence.}
	\label{fig:fig12}
\end{figure}

\subsubsection{Effect of difference-guided adaptation on cross-domain change representations}
\begin{figure*}[!h]
	\centering
	\includegraphics[width=0.8\textwidth]{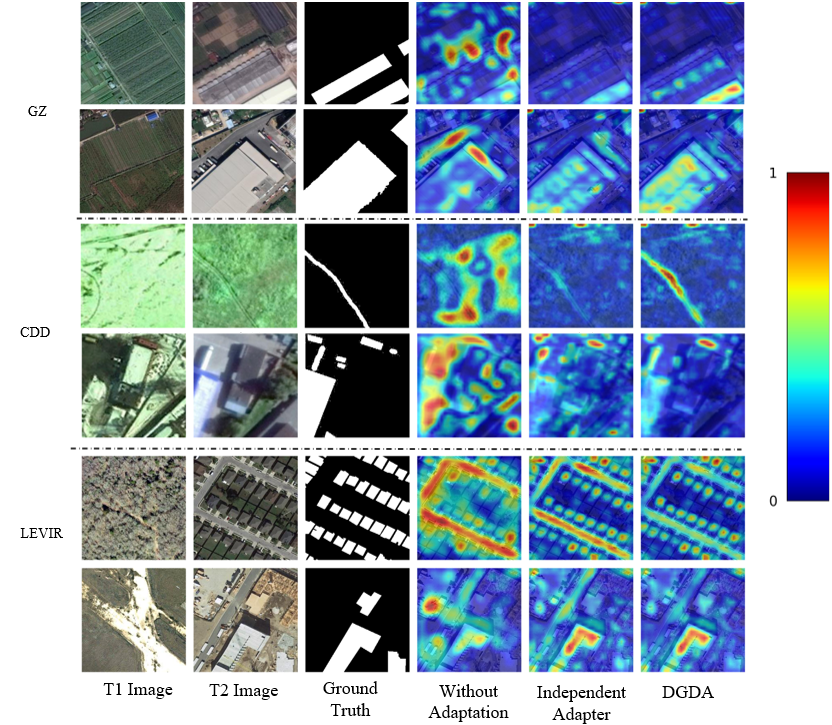}
	\caption{Visualization of channel-averaged feature responses under different adaptation strategies.}
	\label{fig:feature_vis}
\end{figure*}

This section evaluates the impact of different adaptation strategies on change representations under domain shifts. Three configurations are compared: the frozen backbone without adaptation, a parameter-isolation baseline with independent adapters, and the proposed DGDA module. Fig.~\ref{fig:feature_vis} visualizes the channel-averaged activation maps from Layer 3 using representative samples from three datasets.
As shown in Fig.~\ref{fig:feature_vis}, the backbone without adaptation produces dispersed activations dominated by background regions under domain shifts. The responses extend to non-change areas, indicating that the model is sensitive to domain-specific appearance variations rather than discriminative change semantics. Introducing independent adapters partially improves feature adaptation by enabling task-specific adjustment. However, without explicit bi-temporal difference guidance, the model remains unable to effectively separate true changes from temporal appearance variations, resulting in residual activations around irrelevant structures, such as road textures in LEVIR-CD and background patterns in CDD.
In contrast, DGDA produces more compact and change-oriented activation responses across different domains. It suppresses irrelevant background responses while enhancing the spatial concentration of changed objects with clearer structural patterns. This improvement benefits from the explicit integration of bi-temporal difference priors, which guide channel-wise feature recalibration toward change-relevant information. Consequently, DGDA facilitates the disentanglement of domain-specific variations from transferable change semantics, yielding more discriminative representations for DICD.
\subsubsection{Ablation study}
To investigate the contribution of each component, ablation experiments are conducted on the LEVIR $\rightarrow$ CDD $\rightarrow$ GZ sequence, with results summarized in Table~5. The Single-task model serves as the upper bound by independently training each domain and therefore avoids incremental forgetting.
The Baseline (see Appendix A for implementation details), which employs only parameter isolation and conventional knowledge distillation, exhibits noticeable forgetting, with degradation of $-3.47\%$ in $\Delta\text{F1}$ and $-6.27\%$ in $\Delta\text{IoU}$. Adding DGDA to the Baseline reduces the degradation to $-2.63\%$ and $-4.75\%$, respectively, while improving the GZ performance to $86.77\%$ F1 and $76.63\%$ IoU. This improvement demonstrates that difference-guided adaptation effectively enhances change-related representations by suppressing domain-specific interference.
With only FDKD-CS, the degradation is further reduced to $-2.40\%$ in $\Delta\text{F1}$ and $-4.39\%$ in $\Delta\text{IoU}$, achieving $87.45\%$ F1 and $77.70\%$ IoU on GZ. This indicates that frequency-decoupled distillation improves knowledge transfer by maintaining structural consistency and reducing distribution mismatch across domains.
When both DGDA and FDKD-CS are incorporated, the proposed model achieves the best performance, with degradation limited to $-0.49\%$ in $\Delta\text{F1}$ and $-0.99\%$ in $\Delta\text{IoU}$, while obtaining $88.24\%$ F1 and $78.96\%$ IoU on GZ. These results verify the complementary roles of DGDA and FDKD-CS: the former improves domain adaptation through difference-aware representation learning, whereas the latter enhances historical knowledge preservation through frequency-guided distillation.
The qualitative comparisons in Fig.~14 further support these observations. The Baseline produces substantial false responses and missed detections across domains. DGDA improves object completeness in new domains, while FDKD-CS enhances structural consistency and suppresses residual noise. By integrating both modules, the full model generates more coherent predictions across all domains, achieving a favorable stability-plasticity trade-off.
\begin{table}[htbp]
	\centering

	\caption{Ablation study results on the LEVIR $\rightarrow$ CDD $\rightarrow$ GZ incremental sequence. A: Single-task; B: Baseline; C: Baseline+DGDA; D: Baseline+FDKD-CS; E: DG-FDD.}
	\label{tab:ablation_study}
	\begin{tabular}{ccccc}
		\toprule
		\textbf{Setting} & \textbf{Step 1 LEVIR} & \textbf{Step 2 CDD} & \textbf{Step 3 GZ} & \textbf{\boldmath$\Delta$F1 (\%)} \\
		& F1 (\%) \& IoU (\%) & F1 (\%) \& IoU (\%) & F1 (\%) \& IoU (\%) & \textbf{\& \boldmath$\Delta$IoU (\%)} \\
		\midrule
		A & $90.45$ \& $82.57$ & $97.83$ \& $95.75$ & $87.03$ \& $77.04$ & -- \\
		B & $88.06$ \& $78.67$ & $93.17$ \& $87.21$ & $84.42$ \& $73.05$ & $-3.47$ \& $-6.27$ \\
		C & $86.03$ \& $75.49$ & $95.19$ \& $90.83$ & $86.77$ \& $76.63$ & $-2.63$ \& $-4.75$ \\
		D & $86.88$ \& $76.81$ & $94.18$ \& $89.00$ & $87.45$ \& $77.70$ & $-2.40$ \& $-4.39$ \\
		E & $90.17$ \& $82.10$ & $95.32$ \& $91.06$ & $88.24$ \& $78.96$ & $-0.49$ \& $-0.99$ \\
		\bottomrule
	\end{tabular}
\end{table}
\begin{figure}[htbp]
	\centering
	\includegraphics[width=0.8\textwidth]{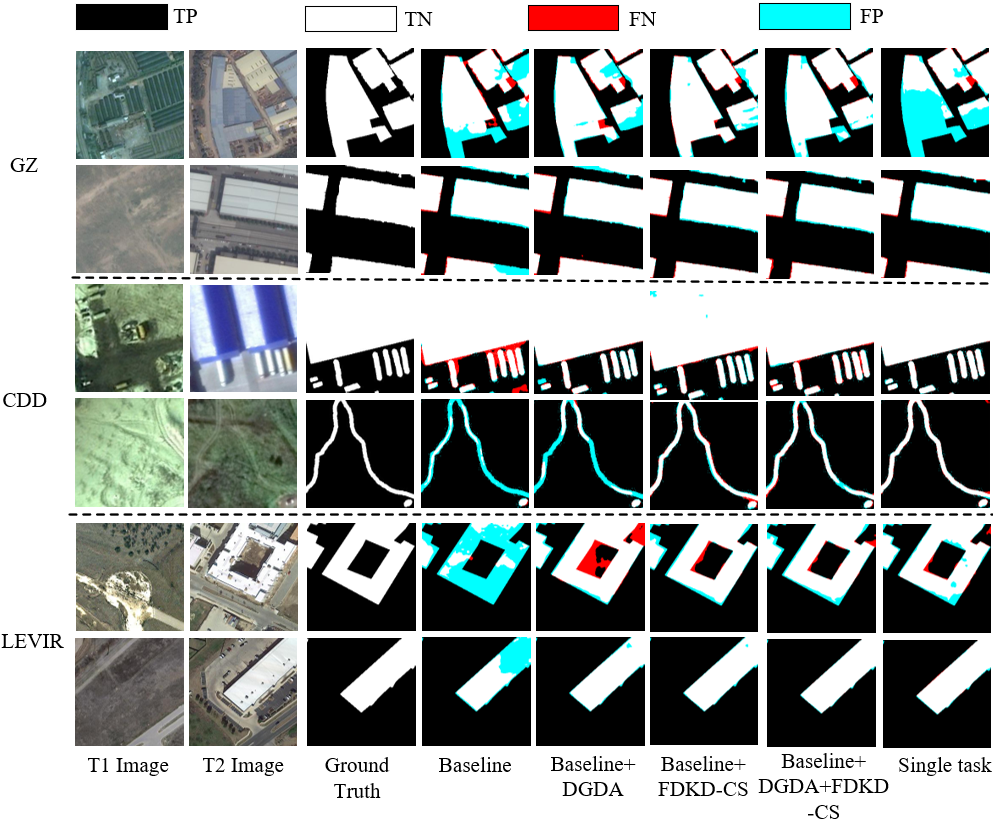}
	\caption{Visualization results of different ablation configurations.}
	\label{fig:fig13}
\end{figure}

\subsection{Efficiency}
To comprehensively evaluate the computational efficiency of DG-FDD during inference, three commonly used metrics are considered: the number of parameters (Parameters), floating-point operations (FLOPs), and per-image inference latency (Time). All methods are evaluated under the same input resolution of $256 \times 256 \times 3$, and the inference time is averaged over 100 random forward passes for reliable comparison. As reported in Table~6, different methods exhibit distinct trade-offs among model complexity, computational cost, and inference efficiency.
\begin{table}[htbp]
	\centering
	\caption{Comparison of model efficiency}
	\label{tab:model_efficiency}
	\begin{tabular}{lccc}
		\toprule
		\textbf{Methods} & \textbf{Model params (M)} & \textbf{FLOPs (Gmac)} & \textbf{Time (ms)} \\
		\midrule
		BIT & $12.40$ & $10.63$ & $19.94$ \\
		STRobustNet & $13.73$ & $20.99$ & $19.43$ \\
		FT & $59.63$ & $115.96$ & $35.53$ \\
		FE & $59.63$ & $115.96$ & $34.66$ \\
		MDIL & $5.27$ & $13.07$ & $98.70$ \\
		MDINet & $41.72$ & $74.04$ & $26.87$ \\
		FOCUS & $41.81$ & $74.27$ & $76.52$ \\
		DI-FRCNN & $41.81$ & $74.27$ & $22.23$ \\
		\midrule
		DG-FDD & $60.15$ & $115.98$ & $42.25$ \\
		\bottomrule
	\end{tabular}
\end{table}
BIT and STRobustNet employ lightweight Transformer-based architectures, resulting in relatively low computational overhead. However, their limited adaptation mechanisms make them less effective in preserving historical knowledge under cross-domain incremental learning. MDIL reduces the model size to $5.27\,\text{M}$ parameters through parameter isolation, but introduces additional inference latency ($98.70\,\text{ms}$) due to its dynamic routing strategy, which requires sequential decision operations.
FOCUS and DI-FRCNN are both implemented on the same ResNet-50 backbone for fair comparison. Therefore, they share identical parameter counts ($41.81\,\text{M}$) and computational costs ($74.27\,\text{GMACs}$). However, FOCUS introduces additional channel-consistency masking and feature reconstruction operations during inference, resulting in higher latency ($76.52\,\text{ms}$) compared with DI-FRCNN ($22.23\,\text{ms}$).
Compared with these methods, DG-FDD introduces moderate computational overhead, with $60.15\,\text{M}$ parameters, $115.98\,\text{GMACs}$, and $42.25\,\text{ms}$ inference time. The additional cost mainly originates from the DGDA module and the multi-head decoder architecture, which introduce difference-guided feature recalibration and task-specific parameter storage, respectively. Nevertheless, during inference, only the shared backbone and the decoder corresponding to the current domain are activated, avoiding unnecessary computational redundancy.
Combined with the performance results in Tables~1 and~2, DG-FDD achieves substantially improved knowledge retention and new-domain adaptation compared with FT while maintaining acceptable computational complexity. These results demonstrate that the proposed difference-guided adaptation and frequency-decoupled distillation mechanisms provide an effective stability--plasticity trade-off with practical inference efficiency, making DG-FDD suitable for domain-incremental change detection.
\section{Conclusion}
To address  catastrophic forgetting in high-resolution RSCD under continual domain shifts, this paper proposed DG-FDD, a domain-incremental change detection framework that integrates difference-guided dynamic adaptation and frequency-decoupled distillation. By combining parameter-efficient adaptation with frequency-domain knowledge regularization, DG-FDD enables effective adaptation to new domains while preserving historical knowledge without accessing previous data. Experiments on LEVIR-CD, CDD, and GZ-CD show that DG-FDD consistently outperforms state-of-the-art methods under both two-task and three-task incremental settings, effectively mitigating catastrophic forgetting while maintaining strong performance on new domains. In several cross-domain scenarios, the method further achieves positive transfer, demonstrating its robustness for continual cross-domain RSCD. Future work will extend the framework to more challenging conditions, including severe class imbalance, multimodal optical--SAR change detection, and long-term large-scale domain shifts.

\appendix

\section{Knowledge confusion and distillation failure in existing incremental methods}

To alleviate catastrophic forgetting, existing DICD methods typically integrate parameter isolation with knowledge distillation strategies for optimization. Parameter isolation methods generally freeze the shared backbone parameters $\theta_{\text{shared}}$, while introducing a small set of domain-specific parameters $\theta_{\text{specific}}^t$ at the current stage $t$, so as to reduce interference from new-domain learning on historical knowledge. However, most of these approaches are derived from single-image classification tasks, and their adaptation modules tend to apply independent and identical feature refinement operations to bi-temporal inputs $X^{t_1}$ and $X^{t_2}$:
\begin{equation}
	\tilde{F}^{t_1} = \mathcal{A}(F^{t_1}; \theta_{\text{specific}}^t)
\end{equation}
\begin{equation}
	\tilde{F}^{t_2} = \mathcal{A}(F^{t_2}; \theta_{\text{specific}}^t)
\end{equation}
where $\mathcal{A}(\cdot)$ denotes the adaptation operator; $F^{t_1}$ and $F^{t_2}$ represent deep feature embeddings extracted from the frozen shared backbone for the two temporal inputs, respectively; and $\tilde{F}^{t_1}$, $\tilde{F}^{t_2}$ denote the recalibrated features after adaptation.

Such independent temporal modeling lacks explicit bi-temporal interaction, and the learned $\theta_{\text{specific}}^t$ in single-stream adapters tends to overfit local appearance patterns of individual images rather than true change semantics. As a result, knowledge confusion is exacerbated under cross-domain scenarios. This highlights the necessity of constructing a parameter-decoupled mechanism that explicitly exploits bi-temporal difference priors, thereby improving the stability of change representation learning.

Another mainstream paradigm adopts knowledge distillation as a functional regularization strategy, where the output distribution of the current model $\Phi_t$ is constrained to match that of the frozen previous model $\Phi_{t-1}$, thereby transferring historical knowledge:
\begin{equation}
	\mathcal{L}_{\text{KD}}^{\text{trad}} = \frac{1}{H \times W} \sum_{h,w} \text{KL}\big(P_{t-1}^{(h,w)}(X_t) \parallel P_t^{(h,w)}(X_t)\big)
\end{equation}
where $\text{KL}(\cdot \parallel \cdot)$ denotes the Kullback--Leibler divergence measuring distribution discrepancy; $P_t^{(h,w)}(X_t)$ represents the predicted probability distribution at spatial location $(h, w)$ of the current model for input $X_t$; and $P_{t-1}^{(h,w)}(X_t)$ corresponds to the frozen previous model.

However, conventional distillation assumes a certain degree of overlap between old and new domain distributions, such that $\Phi_{t-1}$ can still produce meaningful soft labels for $X_t$. In DICD scenarios, severe domain shift leads to a significant deviation of $X_t$ from the training distribution of the old model, and thus $P_{t-1}(X_t)$ becomes highly uncertain:
\begin{equation}
	P_{t-1}(X_t) \approx \text{Noise}, \quad \text{when } \mathbb{P}(X_t) \gg \mathbb{P}(X_{t-1})
\end{equation}

Under this condition, directly applying distillation introduces noisy supervisory signals into optimization, resulting in semantic drift and degraded knowledge transfer, thereby weakening the effectiveness of the constraint. Consequently, under limited access to historical data, it becomes necessary to design a feature transmission mechanism capable of suppressing domain-style interference and reconstructing the old-domain feature distribution. This forms the primary motivation for introducing frequency-domain decoupling and cross-domain pseudo-distribution synthesis in this work.

\subsection*{CRediT authorship contribution statement}

\noindent \textbf{Daifeng Peng}: Writing -- review \& editing, Writing -- original draft, Supervision, Funding acquisition, Formal analysis, Conceptualization.
\textbf{Yaning Li}: Writing --original draft\& editing, Visualization, Methodology, Software, Conceptualization.
\textbf{Haiyan Guan}: Writing --review\&editing, Visualization, Formal analysis.

\subsection*{Declaration of competing interest}

The authors declare that they have no known competing financial interests or personal relationships that could have appeared to influence the work reported in this paper.

\subsection*{Acknowledgements}

This work was supported the National Natural Science Foundation of China (42371449, 41801386).

\bibliographystyle{elsarticle-harv}
\bibliography{references}

\end{document}